\documentclass{article}
\usepackage[preprint]{neurips_2026}
\setcitestyle{numbers,square}
\usepackage[utf8]{inputenc} 
\usepackage[T1]{fontenc}    
\usepackage{hyperref}       
\usepackage{url}            
\usepackage{booktabs}       
\usepackage{amsfonts}       
\usepackage{nicefrac}       
\usepackage{microtype}      
\usepackage{xcolor}         
\usepackage{amsmath}
\usepackage{multirow}
\usepackage{graphicx}
\usepackage{pgfplots}
\usepackage{pgfplotstable}
\usepackage{caption}
\usepackage{tikz}
\usetikzlibrary{positioning,fit,arrows.meta,calc}
\pgfplotsset{compat=1.18}

\title{Mach-Mind-4-Flash Technical Report}

\author{%
  Foundation Model, Li Auto Inc.\\
}

\begin{document}
\maketitle
\begin{abstract}
We present \textbf{Mach-Mind-4-Flash}, a 35B-parameter Mixture-of-Experts (MoE) agentic model with 3B activated parameters. Through post-training optimization alone without scaling pre-training compute, the model achieves performance on par with or surpassing that of 100B-parameter-class models. By introducing scalable agentic interaction environments for large-scale reinforcement learning, the model attains significant performance gains on real-world application tasks. Our pipeline comprises three stages: (1) a unified RL/OPD training infrastructure with dynamic multi-teacher scheduling and operator-level acceleration, delivering 17\% end-to-end training speedup; (2) multiple domain-specific RL experts trained in parallel across Reasoning, General, and Agent tracks, then fused into a single generalist via Multi-Teacher On-Policy Distillation (MOPD)---a routed reverse-KL objective that eliminates the see-saw degradation of mixed-reward RL; (3) Hybrid Median-length Policy Optimization (HMPO), a single-stage token-efficiency method that compresses reasoning chains by 19--46\% with $\le$0.7 percentage-point accuracy loss. Mach-Mind-4-Flash scores 92.70 on AIME'26, 82.82 on IFBench, 80.74 on Behavioral-SafetyBench, 75.80 on BFCL-v4, 72.31 on BrowseComp-zh, and 84.20 on ClawBench---leading or matching models with 10--30$\times$ its activated size at a fraction of the inference cost.
\end{abstract}

\begin{figure}[h]
    \centering
    \includegraphics[width=1.0\columnwidth]{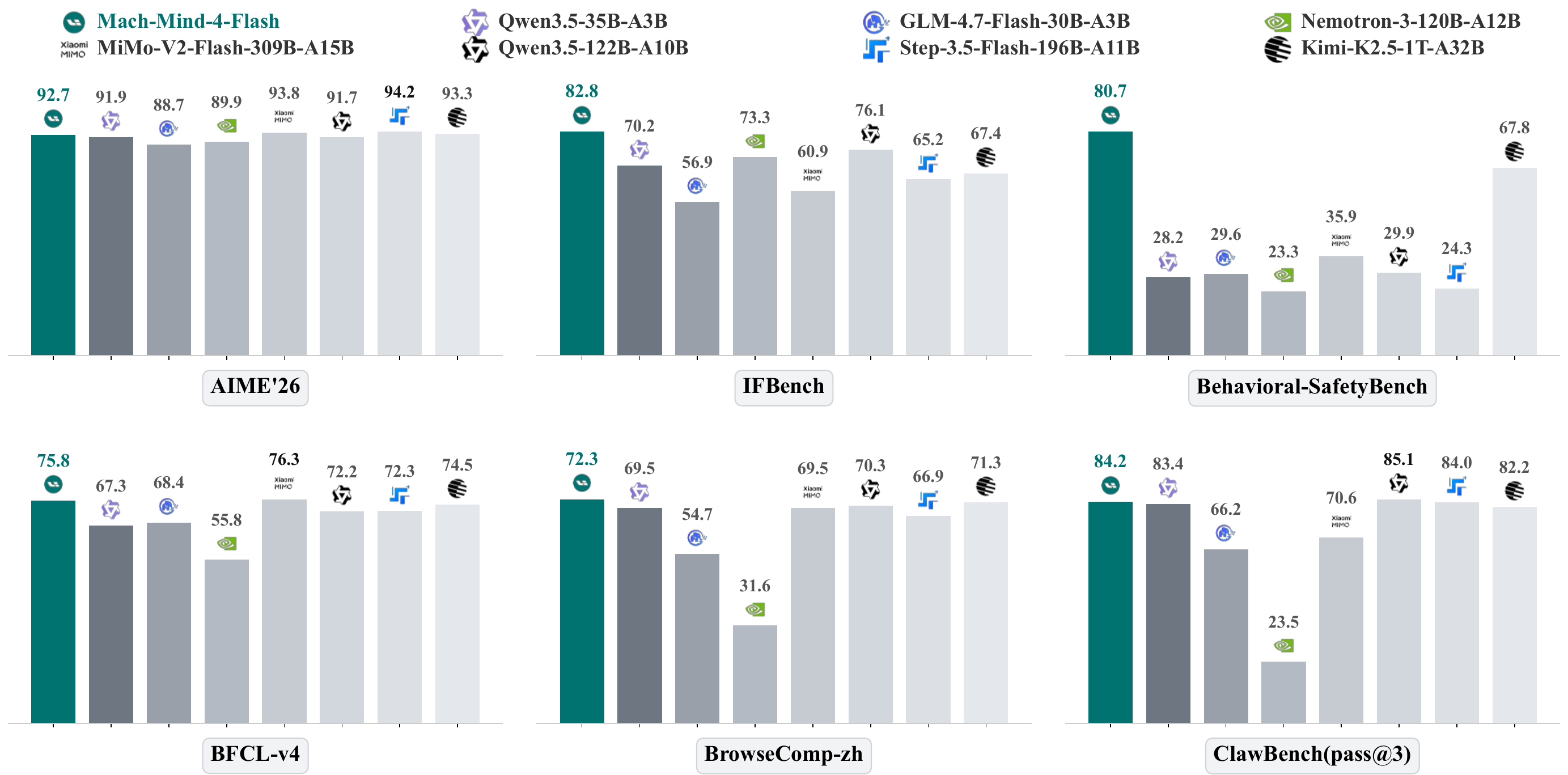} 
    \caption{\textbf{Mach-Mind-4-Flash matches or exceeds much larger models across diverse capability axes.} With only 3B activated parameters, Mach-Mind-4-Flash leads on IFBench, Behavioral-SafetyBench, and BrowseComp-zh, while remaining competitive on reasoning, tool use, and agentic coding against models with 3--30$\times$ its activated size.}
    \label{fig:eval_results}
\end{figure}

\newpage
\tableofcontents
\newpage

\section{Introduction}
Scaling language models has been the dominant recipe for capability gains, yet the associated inference cost makes trillion-parameter models impractical for latency-sensitive deployment. An emerging alternative is to scale the \emph{post-training} pipeline, pushing a compact base model toward frontier performance through reinforcement learning, expert fusion, and inference-time efficiency optimization rather than scaling pre-training compute alone. Mach-Mind-4-Flash is our contribution along this axis: a 35B MoE model (3B activated) whose post-training stack elevates it to the performance tier of models with 10--30$\times$ more activated parameters. Three technical pillars underpin this result:

\paragraph{(1) Scalable training infrastructure with operator-level acceleration.}
Post-training a generalist model requires orchestrating multiple RL expert tracks, a multi-teacher distillation stage, and a token-efficiency stage, all sharing the same distributed infrastructure. We build a unified Reinforcement Learning / On-Policy Distillation (RL/OPD) training paradigm that seamlessly switches between pure RL, pure distillation, and joint modes through a single weighted loss. On top of this, we design a dynamic multi-teacher architecture where adding a new expert requires zero intrusion into the training core. At the operator level, we deeply integrate SonicMoE indexed Grouped General Matrix Multiplication (GEMM) kernels and a segmented shared-expert fusion strategy that overlaps communication with computation, delivering up to 17\% end-to-end speedup on Qwen3.5-35B-A3B training.

\paragraph{(2) Specialization-then-integration: multiple experts, one generalist.}
Rather than training a single model on a heterogeneous reward mixture, which invariably suffers from capability see-saw, we independently train multiple RL experts across three tracks: Reasoning (Math, Code, STEM), General (Instruction Following, Writing, Safety), and Agent (Tool-Use, DeepSearch, Code Agent, Claw Agent). Each expert is optimized with domain-appropriate data synthesis, verifiable reward signals, and tailored training strategies (e.g., two-stage reward curricula for sparse-outcome domains, executable multi-turn environments for agent domains). These specialists are then consolidated through Multi-Teacher On-Policy Distillation (MOPD), which routes each training sample to its corresponding frozen teacher and supervises the student on its own rollouts via a token-level reverse-KL objective. This design parallelizes expert development, eliminates sequential dependencies, and produces a unified model that retains, and occasionally exceeds, individual expert performance.

\paragraph{(3) Token efficiency without accuracy sacrifice.}
Strong reasoning models tend to overthink, producing verbose chains that inflate serving cost without proportional accuracy gains. We address this with HMPO (Hybrid Median-length Policy Optimization), a single-stage RL method applied as the final pipeline step. HMPO derives a group-adaptive length budget from the median of correct rollouts and applies a multiplicative correctness-first reward that mathematically prevents reward hacking on short-but-wrong outputs. Trained only on mathematics, the compression generalizes across code, science, and instruction following, reducing generation length by 19--46\% with at most 0.7pp accuracy loss.

\paragraph{Results.}
The resulting model, Mach-Mind-4-Flash, demonstrates that aggressive post-training can close the gap between a compact 35B MoE and substantially larger frontier models. On competition mathematics (AIME'26: 92.70), code generation (LiveCodeBench-V6: 80.91), instruction following (IFEval: 94.64; IFBench: 82.82), safety (Content-SafetyBench: 98.20; Behavioral-SafetyBench: 80.74), and autonomous agent tasks (ClawBench: 84.20; BFCL-v4: 75.80), Mach-Mind-4-Flash consistently matches or outperforms 120B-class models and remains competitive with trillion-parameter systems---while requiring only 3B activated parameters at inference time. We release this technical report to provide a detailed account of the training methodology, infrastructure innovations, and evaluation results.

\section{Related Work}

Recent large language models are moving from general-purpose chat assistants toward production-oriented agentic systems that can reason, call tools, execute code, interact with environments, and complete long-horizon tasks. We summarize related work from three aspects: agentic foundation models and benchmarks, agentic post-training and reinforcement learning, and efficient and safe deployment.

\subsection{Agentic Foundation Models and Benchmarks}

Early language-agent work showed that LLMs can act through external tools and environments rather than only generate text. ReAct~\cite{yao2023react} interleaves reasoning and actions, while Toolformer~\cite{schick2023toolformer}, Gorilla~\cite{patil2023gorilla}, ToolBench~\cite{qin2023toolbench}, and API-Bank~\cite{li2023apibank} study tool learning, API selection, and tool-augmented instruction following. These works establish the basic agent loop of selecting tools, generating arguments, observing results, and continuing reasoning.

Recent technical reports further frame foundation models as agentic systems. Kimi K2~\cite{kimi2025k2} emphasizes open agentic intelligence with strong reasoning, coding, and tool-use capabilities. GLM-5~\cite{glm5team2026glm5vibecodingagentic} describes the transition from ``vibe coding'' to agentic engineering, targeting complex software engineering and long-horizon workflows. Step-3.5-Flash~\cite{huang2026step35flash} and MiMo-V2-Flash~\cite{xiao2026mimov2flash} highlight efficient sparse activation for low-cost agent deployment, while Qwen3-Coder-Next~\cite{cao2026qwen3codernext} focuses on coding agents trained with executable environments. Meanwhile, agent benchmarks are also moving from static QA to realistic task completion: BFCL~\cite{patil2024bfcl} evaluates function calling, $\tau$-bench~\cite{yao2024taubench} evaluates tool-agent-user interaction, ToolSandbox~\cite{lu2025toolsandbox} studies stateful tool use, and SWE-bench~\cite{jimenez2023swebench} evaluates real software issue resolution. These works suggest that agentic models should be judged by task success, tool reliability, and long-horizon execution rather than final-answer accuracy alone.

\subsection{Agentic Post-training and Reinforcement Learning}

Post-training has become a major driver of model capability. InstructGPT~\cite{ouyang2022training} established the effectiveness of SFT and RLHF for instruction following. More recent reasoning models show that reinforcement learning can further elicit long-chain reasoning and self-improvement. DeepSeek-R1~\cite{guo2025deepseek} demonstrates that RL can induce long chain-of-thought reasoning, verification, and self-reflection. Kimi k1.5~\cite{kimi2025k15}, DAPO~\cite{yu2025dapoopensourcellmreinforcement}, and Skywork-OR1~\cite{he2025skyworkor1} further explore scalable RL recipes, rule-based rewards, and stable large-scale optimization.

For agentic tasks, post-training must handle multi-turn interaction, delayed feedback, sparse rewards, and error recovery. Multi-turn RL for tool-calling agents~\cite{modecrua2026multiturntoolrl} studies reward calibration for intermediate tool actions, while long-context software-engineering-agent training~\cite{golubev2025swerl} shows the value of interactive RL for repository-level tasks. In addition, preference optimization methods such as DPO~\cite{rafailov2023dpo}, ORPO~\cite{hong2024orpo}, SimPO~\cite{meng2024simpo}, and multi-objective DPO~\cite{zhou2024modpo} provide practical alternatives for aligning models with preference signals. These methods are closely related to OPD/MOPD-style training, where multiple objectives such as task success, tool efficiency, token cost, and safety must be optimized jointly instead of being collapsed into a single reward.

\subsection{Efficient and Safe Agent Deployment}

Production agents require not only strong reasoning but also efficient and safe execution. Agent workflows often contain tool schemas, retrieved documents, execution traces, conversation history, and intermediate observations, making long-context inference and prefill cost major bottlenecks. Prompt compression methods such as LLMLingua~\cite{jiang2023llmlingua}, LongLLMLingua~\cite{jiang2023longllmlingua}, and LLMLingua-2~\cite{pan2024llmlingua2} reduce token usage while preserving task-relevant information. DeepSeek-V4~\cite{deepseek2026v4} targets million-token context efficiency, and MiniMax-M1~\cite{chen2025minimaxm1} uses lightning attention to scale test-time compute. At the serving level, vLLM~\cite{kwon2023vllm}, DistServe~\cite{zhong2024distserve}, and multi-token prediction~\cite{gloeckle2024mtp} improve KV-cache management, prefill/decode efficiency, and decoding speed.

Safety is equally important because agents can execute actions rather than merely output text. Constitutional AI~\cite{bai2022constitutional} and Rule-Based Rewards~\cite{mu2024rulebasedrewards} provide principle-based and rule-based alignment mechanisms, while Llama Guard~\cite{inan2023llamaguard} supports safety classification. For deployed agents, safety must cover both content safety and behavior safety, including whether the model calls tools appropriately, respects constraints, and avoids unsafe operations. Therefore, practical agent systems require joint optimization of capability, latency, token cost, and safety.

\section{Infra}

To enhance the training efficiency and system flexibility of large language models (LLMs) during the post-training phase (which encompasses reinforcement learning (RL) and knowledge distillation), we have systematically upgraded the existing mature RL training framework. We propose a unified joint training paradigm for RL and On-Policy Distillation (OPD), upon which we construct a dynamic multi-teacher scalable architecture. Furthermore, we conduct in-depth training acceleration and optimization from two dimensions: the system framework level and the underlying operator level. The core technical innovations are detailed as follows:

\begin{figure}[htbp]
    \centering
    \includegraphics[width=\columnwidth]{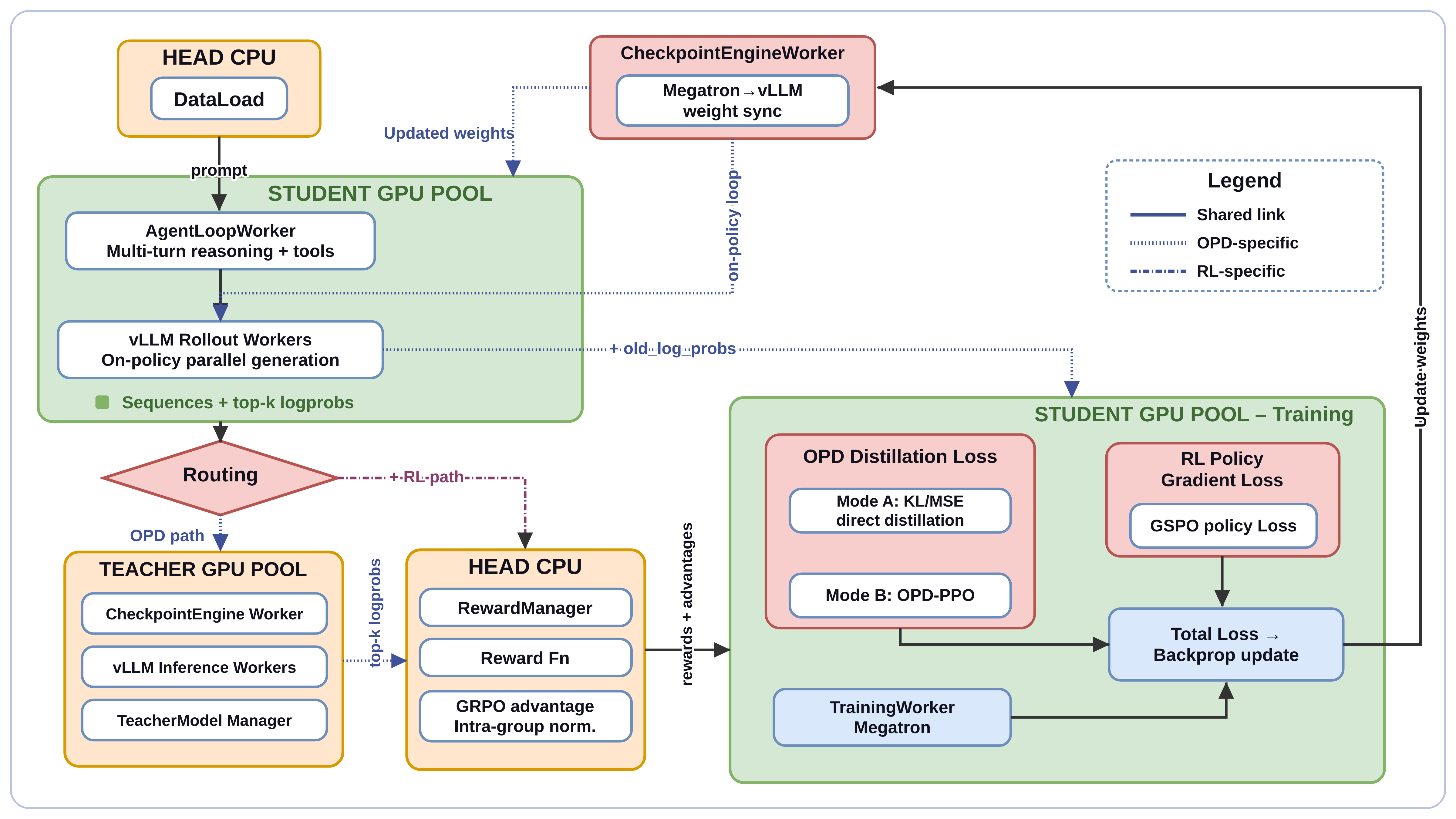} 
    \caption{Unified training framework for RL and OPD.}
    \label{fig:unified_framework}
\end{figure}

\subsection{Unified OPD Training Paradigm Based on the RL Framework}

Traditional knowledge distillation is typically independent of the RL training pipeline, making it difficult to fully leverage the system advantages of the RL framework in distributed scheduling, experience sampling, and policy optimization. To this end, we deeply integrate OPD into the RL framework to construct a unified joint training paradigm. The specific framework is illustrated in Figure~\ref{fig:unified_framework}.

This paradigm achieves global optimization control through a unified weighted loss function:
\begin{equation}
    \mathcal{L} = \alpha \cdot \mathcal{L}_{\text{OPD}} + \beta \cdot \mathcal{L}_{\text{RL}}
\end{equation}
where:
\begin{itemize}
    \item $\mathcal{L}_{\text{OPD}}$: Applied to the distillation path. To accommodate complex knowledge transfer requirements, we extend the loss function forms of OPD. In addition to the traditional MSE loss, it supports various distillation objectives such as Forward\_kl\_TopK, thereby driving the student model's output distribution to approximate the teacher's target distribution more accurately and efficiently.
    \item $\mathcal{L}_{\text{RL}}$: Applied to the policy gradient path. It performs sequence-level clipped updates based on task reward signals.
\end{itemize}

By dynamically adjusting the weight coefficients $\alpha$ and $\beta$, this formulation can flexibly switch between three training modes:
\begin{enumerate}
    \item \textbf{Pure RL Mode ($\alpha = 0, \beta > 0$):} The distillation path is disabled, and the framework degenerates into standard RL training, fully retaining native RL capabilities such as policy gradient updates and multi-turn tool calling.
    \item \textbf{Pure OPD Mode ($\alpha > 0, \beta = 0$):} Policy updates based on task rewards are disabled, retaining only the distillation path. The student model undergoes online distillation entirely using the teacher as the supervision target. This mode exhibits more stable convergence and is suitable for the model capability initialization phase.
    \item \textbf{Joint RL + OPD Mode ($\alpha > 0, \beta > 0$):} Both distillation and policy gradient losses are enabled. The two optimization paths are mutually independent and naturally superimposed, demonstrating excellent compatibility.
\end{enumerate}

Benefiting from this deep integration, the unified paradigm directly inherits the system-level capabilities of the RL framework:
\begin{itemize}
    \item Reuses mature distributed training and scheduling mechanisms, significantly reducing the additional engineering complexity of the distillation system;
    \item Inherits the asynchronous reward routing mechanism, supporting parallel reward computation for over 20 types of tasks;
    \item Achieves a closed-loop implementation of online sampling, reward evaluation, policy optimization, and distillation supervision within a single framework, significantly enhancing the consistency and maintainability of the training pipeline.
\end{itemize}

\subsection{Dynamic Multi-Teacher Scalable Architecture}
\begin{figure}[h]
    \centering
    \includegraphics[width=\columnwidth]{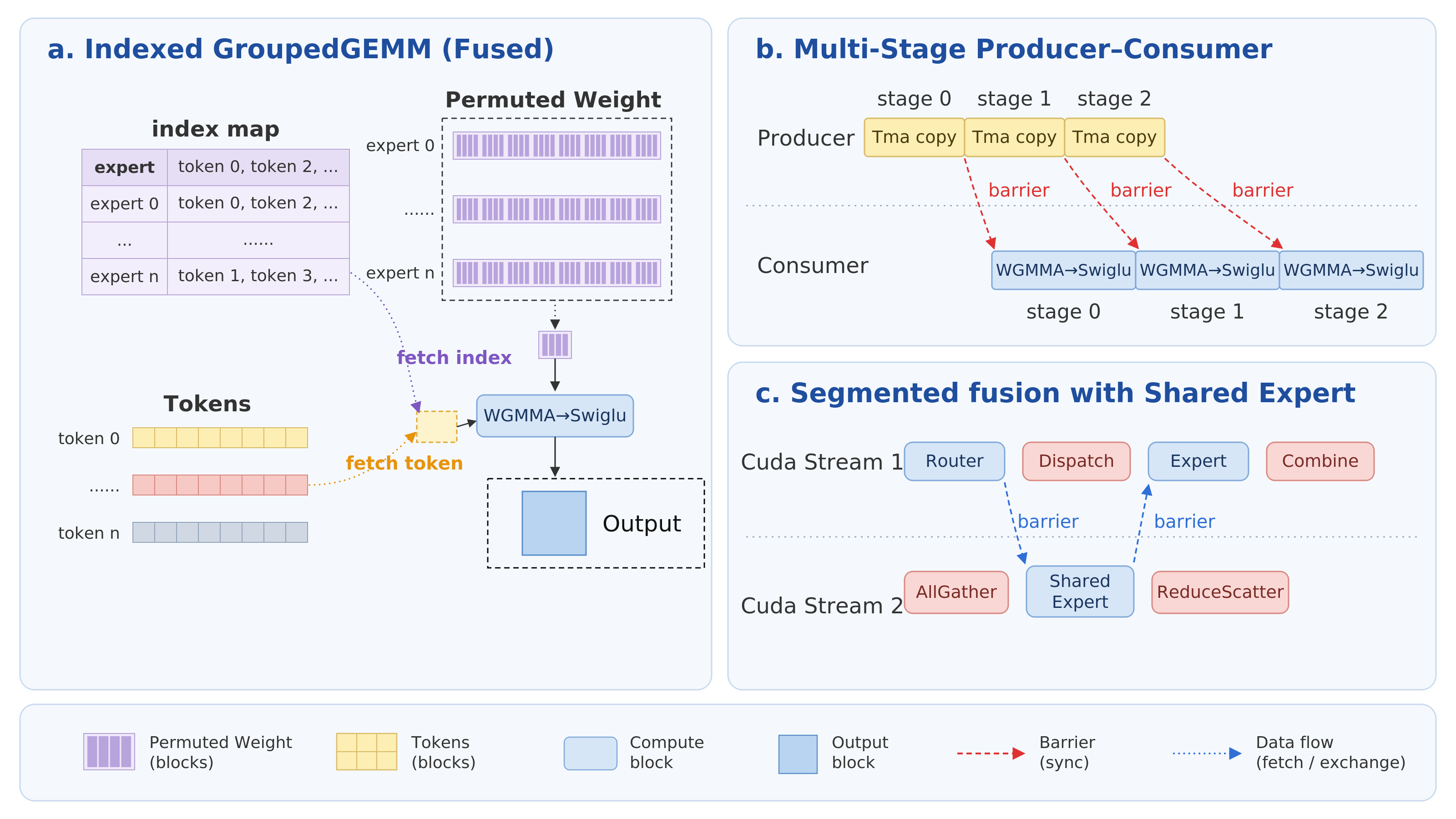} 
    \caption{Underlying operator framework diagram.}
    \label{fig:operator_framework}
\end{figure}

In practical large-scale post-training scenarios, a single teacher model often struggles to comprehensively cover all task domains. Conversely, building independent training pipelines for different tasks easily leads to resource waste and engineering fragmentation. Therefore, we design a dynamic multi-teacher scalable architecture that supports the flexible, on-demand integration of teacher models, enabling a ``building-block'' style combinatorial training capability.

\paragraph{Architecture Design and Routing Mechanism}
The core design principle of the multi-teacher architecture is: adding a new teacher requires zero intrusion into the framework's core logic and remains completely transparent to the student model's training process.

Specifically, each teacher instance is registered as an independent configuration node in the distillation configuration tree. After the Rollout phase is completed, the framework asynchronously distributes samples to the corresponding teacher instances based on the routing identifier of each data entry. After obtaining the target logits, they are uniformly aggregated to compute the distillation loss. This process is completely decoupled from the training updates of the student model, avoiding interference from the teacher-side logic on the main training pipeline. At the data level, simply adding a routing identifier column to the samples enables binding and dynamic switching to any teacher model without modifying the training code throughout the process, endowing the system with extreme flexibility.

\paragraph{Elastic Resource Scheduling}
This architecture relies on the Ray cluster to achieve transparent multi-node resource scheduling. The GPU allocation for the student model (Actor) nodes and teacher nodes is automatically bound by the framework based on the tensor parallelism (TP) scale and the number of replicas, requiring no manual intervention. Each teacher can independently configure its number of replicas, and Ray automatically schedules the corresponding processes to the designated GPUs.

Meanwhile, the parallel inference of multiple teachers and the training of the student model are decoupled via Ray's asynchronous mechanism, ensuring that an increase in the number of teachers does not become a serial bottleneck for training throughput. When business needs require the introduction of a teacher for a new domain, one only needs to request additional nodes and complete the registration to achieve seamless integration through ``configuration-as-scaling.''

\subsection{Training Acceleration and Extreme Optimization}

To achieve extreme throughput in complex joint RL and distillation training scenarios, we conduct deep optimizations from both the system framework level and the underlying operator level.

\paragraph{System and Framework-Level Acceleration}
\begin{itemize}
    \item \textbf{Multi-Dimensional Hybrid Parallelism:} The student model is based on the Megatron-Core backend, natively supporting multi-dimensional parallel strategies such as Tensor Parallelism (TP), Pipeline Parallelism (PP), and Expert Parallelism (EP). Combined with MoE Grouped GEMM operator fusion technology, this significantly improves the training throughput of MoE architecture models.
    \item \textbf{Fine-Grained Operator-Level Recomputation:} A selective activation recomputation strategy is adopted, enabling recomputation only for lightweight operators (e.g., RMSNorm) while skipping compute-intensive operators (e.g., Linear). This strategy strikes an optimal balance between saving GPU memory and controlling computational overhead, allowing the system to support larger batch sizes and further enhancing overall training efficiency.
    \item \textbf{Multi-Token Prediction (MTP) Acceleration:} The framework supports Megatron-Core's native MTP module and utilizes the MTP head for speculative decoding during the vLLM sampling phase, substantially reducing the time consumed in the Rollout phase.
\end{itemize}

\paragraph{Underlying Operator-Level Optimization}

\label{sec:infra-operator-acceleration}

\begin{itemize}
    \item \textbf{Deep Integration of SonicMoE:} Targeting the most compute-intensive MoE MLP module in the Transformer layer, we deeply integrate SonicMoE into the Megatron framework. As shown in Figures~\ref{fig:operator_framework}a and \ref{fig:operator_framework}b, SonicMoE fully utilizes the hardware characteristics of Hopper architecture GPUs (such as the TMA copy engine, Warp specialization, and multi-stage producer-consumer pipelines) to implement an efficient Indexed Grouped GEMM operator and its Gate-Up fused variant. This design eliminates the token permutation step, reduces global memory access latency, and greatly improves GPU computational throughput. During the backward propagation phase, the system introduces a local recomputation strategy to reduce memory access overhead and employs DeepEP instead of traditional All-to-All collective communication, further boosting multi-node communication efficiency.
    \item \textbf{Segmented Fusion with Shared Expert:} As shown in Figure~\ref{fig:operator_framework}c, building upon the introduction of SonicMoE, our efficient segmented fusion strategy for shared experts achieves communication-computation overlap. We introduce a multi-stream mechanism and split the computation of the shared expert into three stages: AllGather, Expert Computation, and ReduceScatter. These are stagger-fused with the computation stages of standard experts under the EP mode (Dispatch, Expert Computation, and Combine), achieving communication-computation overlap and hiding the time consumption of the shared expert. Additionally, we found that when TP, EP, and ETP are enabled simultaneously, if the shared expert only employs TP parallelism, it typically leads to a surge in communication volume. To address this, we forcibly set TP=1 in the shared expert module, reducing the idle waste of computational resources caused by communication. This strategy schedules the shared expert's computation in parallel with the main computation of SonicMoE, effectively masking communication latency and thereby further breaking through the bottleneck of training speed.
\end{itemize}


\section{Post-Training}

\begin{figure}[htbp]
    \centering
    \includegraphics[width=\linewidth]{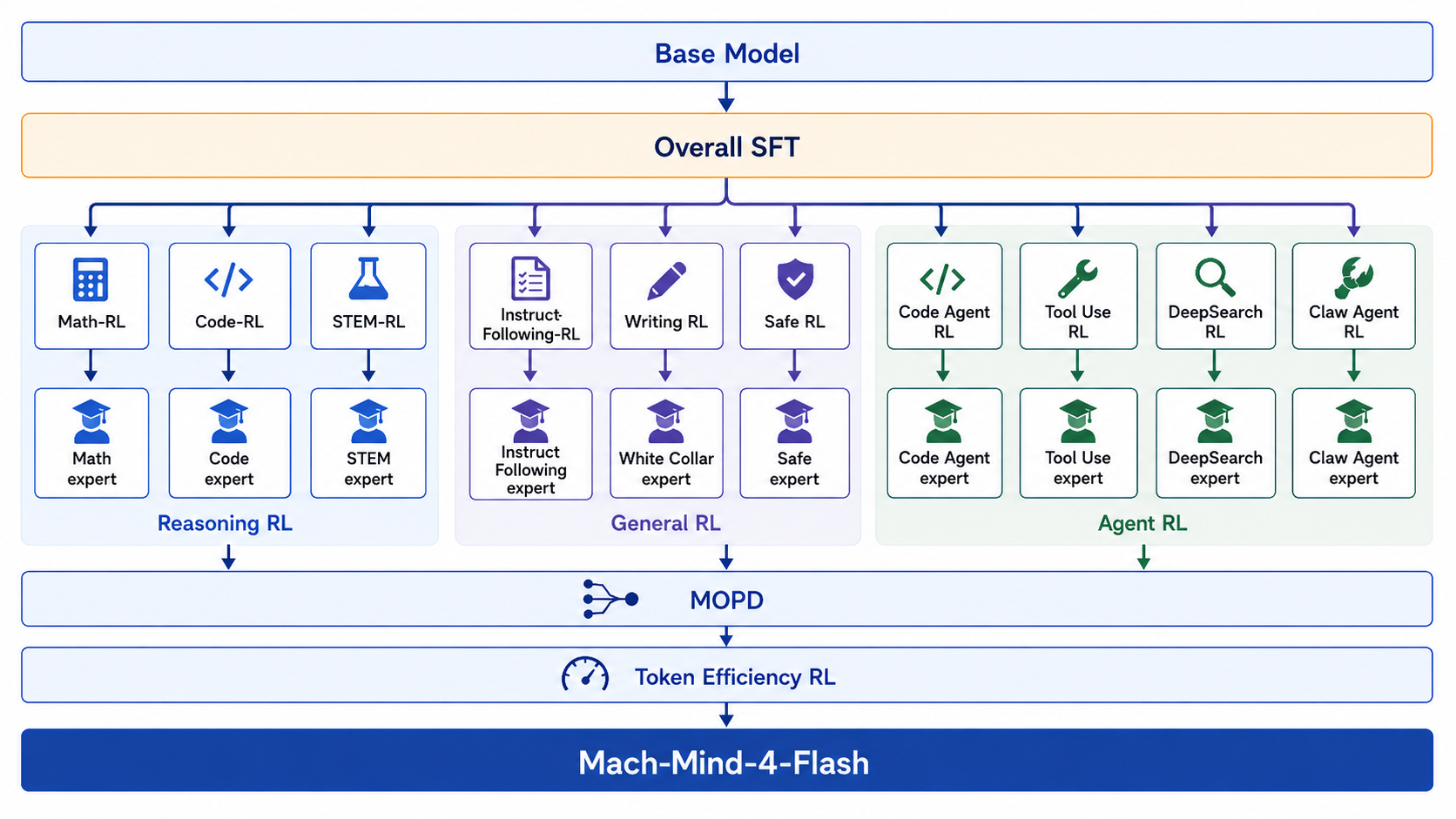}
    \caption{The post-training pipeline of Mach-Mind-4-Flash. The process starts from a base model, followed by overall SFT, domain-specific expert RL training across three parallel tracks, Multi-Teacher On-Policy Distillation (MOPD) for model fusion, and a final token efficiency optimization stage.}
    \label{fig:overall_framework}
\end{figure}

As illustrated in Figure~\ref{fig:overall_framework}, our post-training pipeline for Mach-Mind-4-Flash adopts a hierarchical ``specialization-then-integration'' approach. Building upon Qwen3.5-35B-A3B~\cite{qwen3.5}, we first perform Supervised Fine-Tuning (SFT) to establish foundational alignment. We then diverge the training into three parallel Reinforcement Learning (RL) tracks, each producing multiple specialist checkpoints: \textit{Reasoning RL} (Math, Code, STEM), \textit{General RL} (Instruction Following, Writing, Safety), and \textit{Agent RL} (Tool-Use, DeepSearch, Code Agent, and Claw Agent). To consolidate these diverse experts into a unified model without performance degradation, we introduce Multi-Teacher On-Policy Distillation (MOPD) for seamless expert fusion. Finally, a Token Efficiency RL stage compresses generation length while preserving accuracy, yielding the deployed Mach-Mind-4-Flash.

\subsection{Supervised Fine-Tuning}

To establish a robust foundation for subsequent reinforcement learning, we conduct large-scale multi-domain Supervised Fine-Tuning (SFT) starting from Qwen3.5-35B-A3B~\cite{qwen3.5}. The goal is to align the base model with diverse task formats and equip it with broad capabilities across reasoning, general conversation, and agentic interaction.

\begin{figure}[htbp]
    \centering
    \includegraphics[width=0.8\linewidth]{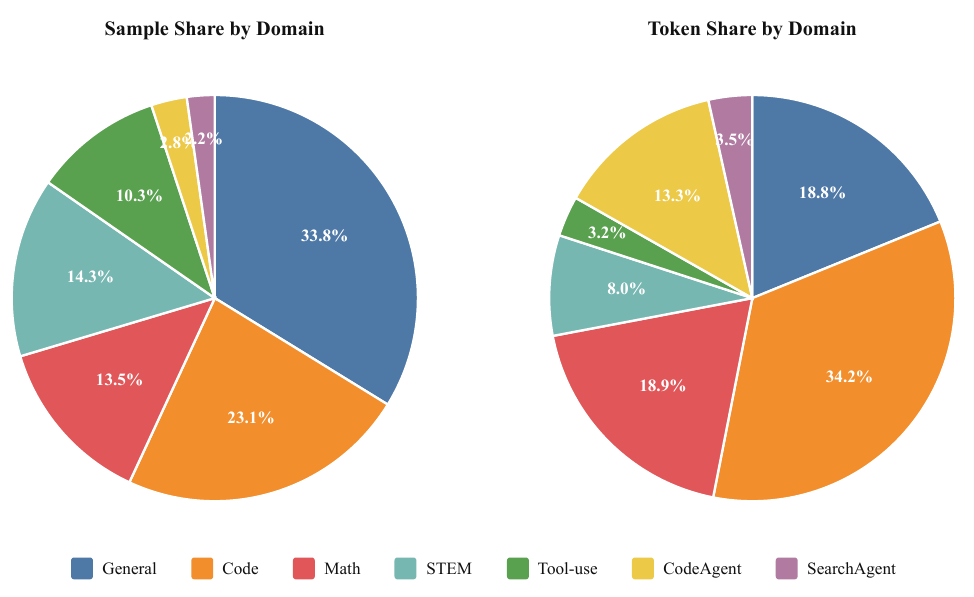}
    \caption{Distribution of our SFT corpus across domains. \textbf{Left:} share of training samples. \textbf{Right:} share of training tokens. General contributes the largest sample fraction, while Code and Math dominate the token budget due to longer reasoning chains. Code Agent exhibits the most pronounced sample-to-token amplification, reflecting the long-horizon nature of agentic trajectories.}
    \label{fig:sft_data}
\end{figure}

\paragraph{Data Composition.}
Our SFT corpus spans seven domains. The distribution of sample counts and token volumes is shown in Figure~\ref{fig:sft_data}. Key design choices for each domain are:

\begin{itemize}
    \item \textbf{Math and STEM:} We focus on verifiable problems with unambiguous ground-truth. High-quality reasoning chains are synthesized via rejection sampling from stronger teacher models, with a difficulty-based filter retaining only samples that challenge the current base model.
    \item \textbf{Code:} Long-form code generation and competitive programming traces, distilled from capable teacher models and filtered by execution-based correctness verification.
    \item \textbf{General:} Covers instruction following, creative writing, and open-ended QA. Response style is optimized for logical clarity and conciseness, with data distilled from stronger teacher models and filtered by multi-dimensional quality rubrics.
    \item \textbf{Tool-use, Code Agent, and DeepSearch:} We construct over 10K diverse environments for agentic trajectory synthesis, focusing on repository-level software engineering and long-horizon multi-turn tasks. Trajectories are generated across multiple scaffolds and filtered by task-level verification.
\end{itemize}

Our data strategy prioritizes \textit{quality} and \textit{reasoning density} over volume. Hard-to-solve samples for the base model are identified and re-annotated by more capable teacher models. Multi-dimensional filtering (instruction following, logical consistency, factual accuracy) is applied through both automated heuristics and human-in-the-loop verification.

\paragraph{Training Configuration.}
We train with a global batch size of 32, a learning rate of $1 \times 10^{-5}$ with cosine decay, and a maximum sequence length of 131,072 tokens. The model is trained for 2 epochs. For agentic trajectories, we apply \textit{error masking}: trajectory segments containing erroneous actions are retained as input context but masked from the loss computation, allowing the model to observe and learn error-recovery behaviors without reinforcing incorrect actions. To improve training throughput, we adopt sample packing that concatenates multiple examples into a single sequence up to the context limit, with attention masks preventing cross-contamination between packed samples.

\subsection{Reasoning and General RL}
We train two parallel RL expert tracks---Reasoning (Math, Code, STEM) and General (Instruction Following, Writing)---that share the same training framework but differ in data sources and reward definitions. Both tracks are trained independently using Group Relative Policy Optimization (GRPO)~\cite{shao2024deepseekmathpushinglimitsmathematical} with domain-specific hyperparameters, each producing a specialist checkpoint from the same SFT initialization that later serves as a frozen teacher in MOPD (Section~\ref{sec:opd-mopd}).

\paragraph{Data Synthesis and Difficulty Pruning.}
For reasoning domains (Math, Code, STEM), we curate tasks where correctness is \emph{deterministically verifiable}---mathematical problems with unique numerical answers, coding problems with executable test suites, and STEM questions with unambiguous ground-truth. We employ LLM-assisted problem synthesis: seed templates are parameterized and transformed to produce diverse novel problems while preserving verifiability. For instruction-following tasks, inspired by the formal constraint grammar of LexInstructEval~\cite{ren2026lexinstructeval}, we decompose instructions into composable $\langle\texttt{Procedure, Relation, Value}\rangle$ triplets and systematically generate constraint-rich prompts through combinatorial composition. Compliance is verified through a two-stage reward: a deterministic programmatic engine first checks each atomic constraint, and samples that pass all rule-based checks are further evaluated by an LLM judge to detect reward hacking (e.g., satisfying constraints via degenerate or nonsensical outputs). Only responses passing both stages receive a positive reward. For writing tasks, we adopt the automated coarse-to-fine rubric generation framework of RubricHub~\cite{li2026rubrichub} to produce comprehensive, highly discriminative evaluation criteria at scale, which serve as both the data synthesis backbone and the reward signal for RL training.

Regardless of domain, we apply unified \textit{Difficulty-based Pruning}. For each candidate query, we perform 8 independent inference passes using the current SFT model, discarding samples that are too trivial (8/8 pass) or too hard (0/8 pass). Retaining only moderate-difficulty samples ensures that the RL training signal remains both informative and learnable.

\begin{figure}[t]
    \centering
    \includegraphics[width=0.8\linewidth]{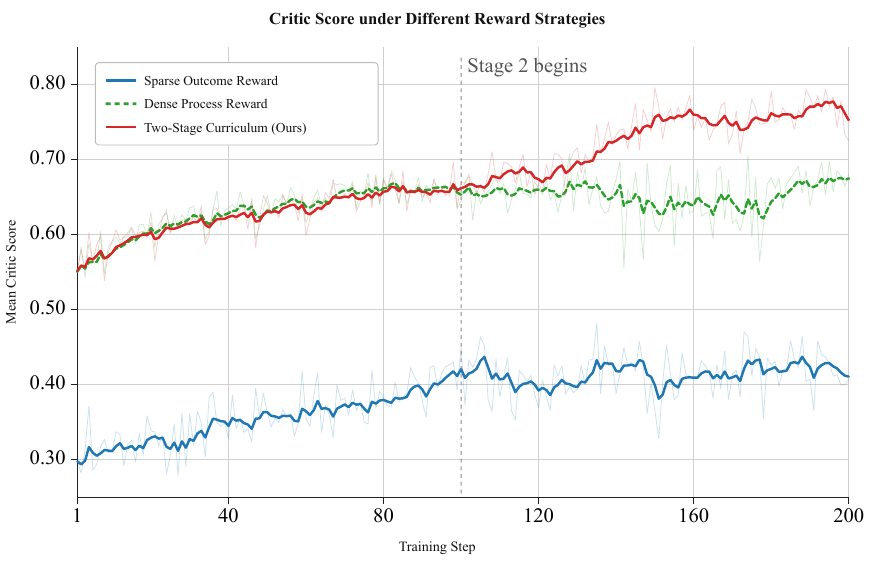}
    \caption{Comparison of reward strategies on a structured reasoning task (Table QA). \textbf{Sparse Outcome Reward} suffers from the zero-reward trap. \textbf{Dense Process Reward} saturates due to reward hacking. Our \textbf{Two-Stage Curriculum} transitions at the marked boundary, combining stable early learning with outcome-aligned improvement. The two-stage schedule is applied selectively to domains where sparse outcome signals impede cold-start convergence.}
    \label{fig:reward_comparison}
\end{figure}

\paragraph{Reward Design.}
Our reward strategy is domain-dependent. For domains where the outcome signal is sparse and the zero-reward trap is severe (e.g., structured reasoning tasks such as Table QA), we employ a \textit{Two-Stage Reward Curriculum}: \textbf{Stage~1} uses a dense process reward $\mathcal{R}_{\text{proc}}$ that provides partial credit for intermediate reasoning steps, enabling stable cold-start learning; once validation performance crosses a predefined threshold, we transition to \textbf{Stage~2}, switching to a strict outcome reward $\mathcal{R}_{\text{out}}$ that forces the model to internalize end-to-end correctness.

For domains where the outcome signal is already informative (e.g., Code with multi-test-case pass rates, Instruction Following with per-constraint compliance), we directly adopt outcome-based rewards without the process scaffolding stage. For writing, the multi-dimensional rubrics generated by RubricHub~\cite{li2026rubrichub} provide fine-grained, criterion-level reward signals that enable effective RL optimization on open-ended generation tasks. The overall principle is consistent: we use the densest verifiable signal that the domain affords, while ensuring the final optimization target is always grounded in task-level correctness. Figure~\ref{fig:reward_comparison} illustrates the benefit of the two-stage curriculum on a structured reasoning task where the zero-reward trap is most pronounced.

\subsection{Safety RL}

Large language models and agentic systems increasingly generate content and execute tool-mediated actions that can affect users, third parties, and the broader information environment, making safety-oriented training essential for reducing harmful outputs and risky agent behavior while preserving general-purpose utility~\cite{bommasani2021foundation,weidinger2021ethical,bender2021dangers}. We construct a safety-oriented Teacher Model that targets two complementary dimensions: \textbf{content safety}, which prevents unsafe or norm-violating responses, and \textbf{behavioral safety}, which prevents unauthorized, irreversible, or instruction-inconsistent tool use. The learned safety behavior is intended to be transferred to downstream models through Multi-Teacher On-Policy Distillation (MOPD), supporting scalable alignment without unnecessary capability loss~\cite{ouyang2022training,bai2022constitutional}.

\paragraph{Data Pipeline.}
The data pipeline combines content-safety data and behavioral-safety data as complementary supervision sources. For content safety, prompts and responses are organized according to a safety taxonomy covering major risk scenarios, value alignment, social norms, public order, and localized requirements, with boundary cases and contrastive examples emphasized to clarify the distinction between safe assistance and unsafe completion~\cite{weidinger2021ethical,gehman2020realtoxicityprompts}. For behavioral safety, tool-use scenarios are constructed around risk level, user authorization, parameter validity, reversibility, and consistency between tool calls and final responses, and pairwise contrastive samples are used so that the model learns when a tool should or should not be invoked~\cite{yao2023react,schick2023toolformer,nakano2021webgpt}. Together, these two data streams improve safe response generation and safe action selection in agentic settings.

\paragraph{Training Pipeline.}

For content safety, we train a dedicated safety Reward Model from the business safety specification through staged SFT, covering specification memorization, boundary-case understanding, and safety judgment~\cite{christiano2017deep,ziegler2019fine,ouyang2022training}. During reinforcement learning, this model assigns low rewards to unsafe responses and optimizes safe responses for usefulness, encouraging helpful restraint while penalizing unsafe completion, policy circumvention, and unnecessary refusals. For behavioral safety, rule-based rewards evaluate the tool-use trajectory based on the safety of tool execution and parameters. Pairwise contrastive samples further distinguish valid tool calls from unsafe, unnecessary, or mismatched actions. Finally, the content and behavioral rewards are then combined to train the safety-oriented Teacher Model, whose safety capability can be distilled into downstream models through MOPD.

\subsection{Tool-Use RL}

\paragraph{Motivation and task formulation.} Tool-Use in production agents is a long-horizon interaction problem rather than an isolated function-call formatting problem. A capable model must ground tool schemas, inspect external state, bind arguments, interpret observations, recover from invalid actions, and decide when the user goal has been completed. Therefore, Tool-Use RL optimizes complete multi-turn trajectories
instead of single-step tool-call accuracy. Given a user goal, an environment state or resource state, an available action/tool set, and task-specific verifiers, the policy repeatedly observes the environment, takes actions, receives feedback, and finally produces a response. The objective is trajectory-level task completion under the same interaction protocol used at inference time. The resulting checkpoint is used as the Tool-Use RL specialist teacher in MOPD.

\begin{figure}[htbp]
    \centering
    \includegraphics[width=\linewidth]{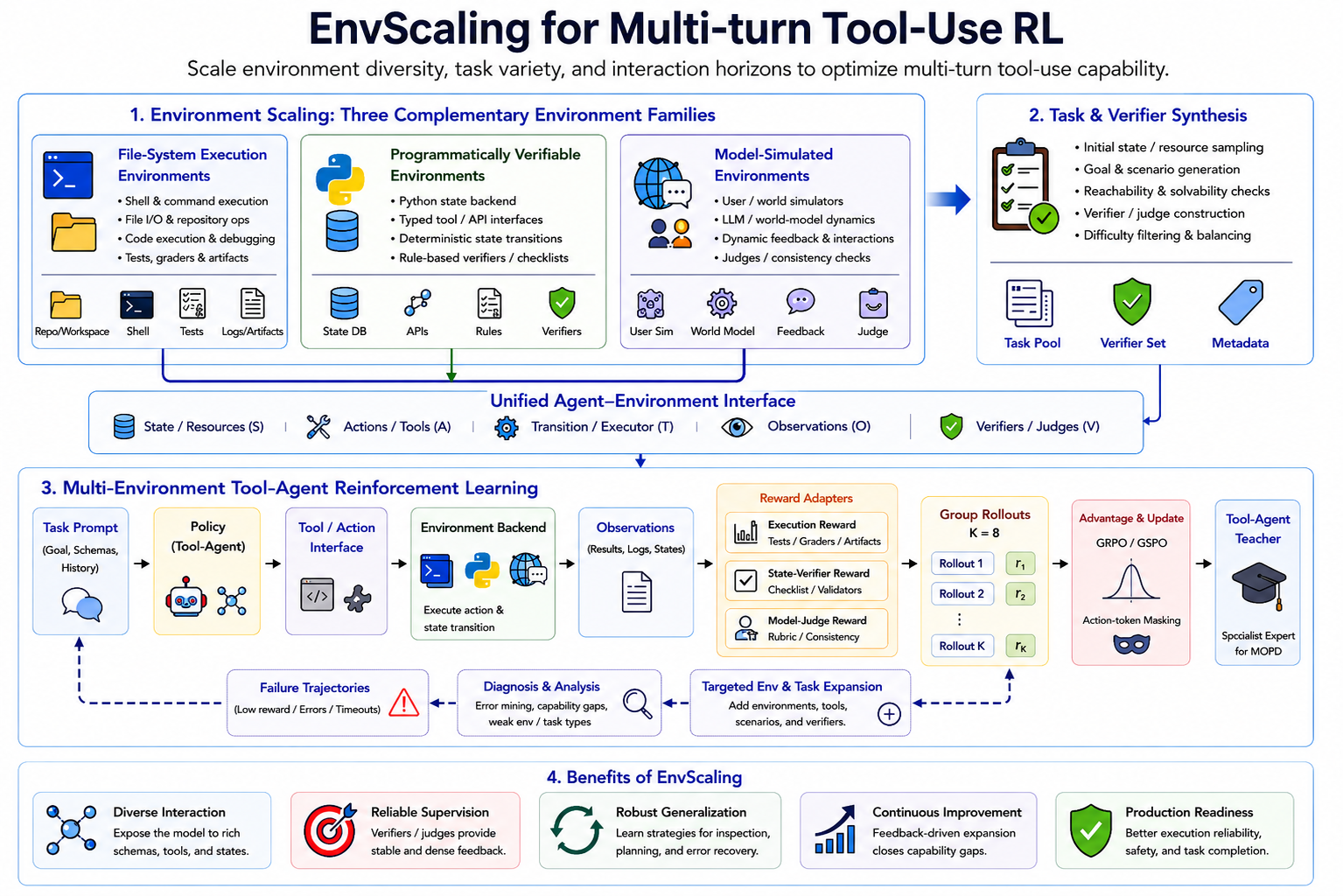}
    \caption{
    Overview of EnvScaling for Tool-Use RL. We scale three complementary environment families: file-system execution environments, programmatically verifiable environments, and model-simulated environments. These environments are unified through a common agent-environment interface and used to synthesize verifiable multi-turn tasks. During RL, executable rollouts are scored by environment-specific reward adapters and optimized with group-relative policy updates under action-token masking.
    }
    \label{fig:envscaling_pipeline}
\end{figure}

\paragraph{EnvScaling and data construction.} We introduce EnvScaling as the main data strategy for Tool-Use RL. Instead of scaling static tool-call traces, we scale the environments in which agent decisions are executed, observed, and verified. Our environment pool contains three complementary families. First, file-system execution environments expose sandboxed workspaces, shell commands, file I/O, code execution, tests, graders, logs, and artifacts, training realistic operational behaviors over persistent resources.
Second, programmatically verifiable environments are Python-simulated stateful environments with typed tool APIs, deterministic transition rules, and state-based validators, providing reliable supervision for schema grounding, state inspection, argument construction, state modification, and error recovery.
Third, model-simulated environments use LLM- or world-model-based simulators to provide user feedback, dynamic world states, and open-ended interaction signals, with judges or rubrics used for evaluation.

All environments are normalized into a unified agent-environment interface consisting of state, actions, transitions or executors, observations, and verifiers or judges. For the
programmatically verifiable family, we use an LLM-assisted generate--build--verify pipeline to synthesize Python state backends, domain constraints, typed tools, and validators. Static checks and positive/negative tool-call tests are applied before rollout. The retained pool covers more than 190 stateful domains and over 3.5K tool interfaces, with each environment typically exposing 10--30 tools. On top of the scaled environment pool, we synthesize multi-turn scenarios by sampling initial states or resources, generating natural-language goals, and filtering candidates by reachability, executability, schema compatibility, and verifier coverage. The tasks are not derived from fixed tool-call scripts; instead, they allow multiple valid solution paths, encouraging information gathering, state tracking, planning, clarification, and recovery.

\paragraph{Reward Design and Training Pipeline.} Tool-Use RL uses trajectory-level rewards because the quality of an action is often revealed only after several later turns. A locally valid tool call may still fail the task, while a trajectory can recover from an invalid action through later inspection and correction. We therefore treat single-step format correctness as a protocol constraint and use end-to-end task completion as the main optimization signal. Different environment families use different reward adapters: file-system environments rely on execution results, tests, graders, and artifact checks; programmatically verifiable environments use final-state checklists and deterministic validators; model-simulated environments use rubric-based judges, simulator-consistency checks, or goal-satisfaction evaluation. Protocol checks over parseability, schema typing, argument validity, tool-execution status, and observation-grounded finalization are used as hard filters or auxiliary signals.
For each prompt, we sample $K=8$ executable trajectories and optimize a group-relative clipped policy objective over their trajectory-level rewards. We use an asymmetric clipping range with
$\epsilon_{\mathrm{low}}=0.20$ and $\epsilon_{\mathrm{high}}=0.28$, bound each trajectory by at most $T_{\max}=40$ assistant/tool-use turns, and disable explicit KL regularization. During optimization, system prompts, user turns, tool observations, environment messages, and padding are retained as context but masked from the policy loss; only assistant-generated decision tokens are optimized. This focuses the RL signal on multi-turn tool-use decisions and improves tool selection, argument binding, observation interpretation, state tracking, error recovery, and final task completion.
\subsection{DeepSearch RL}
\begin{figure}[h]
  \centering
  \includegraphics[width=\textwidth]{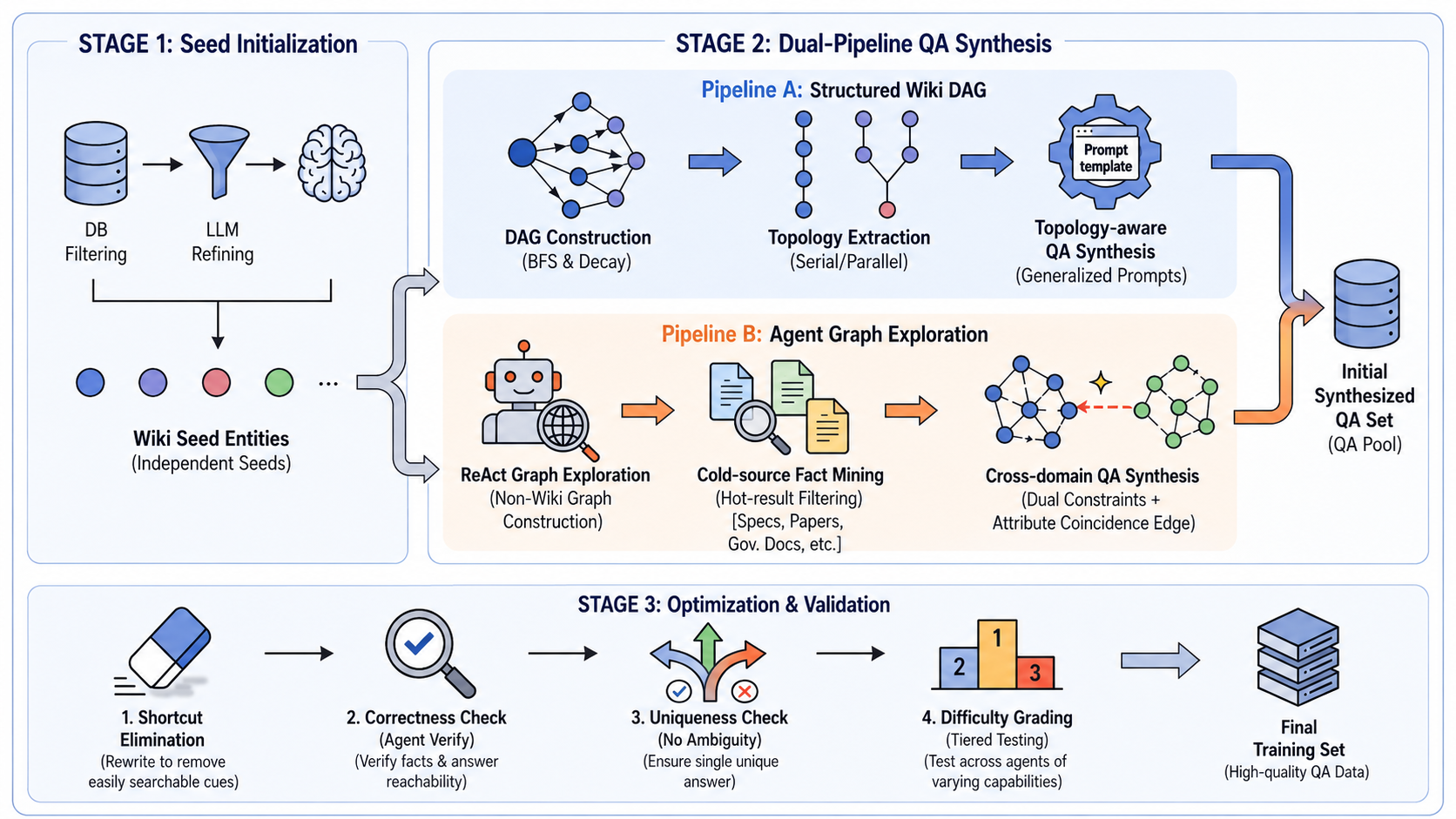}
  \caption{Overview of the DeepSearch High-Difficulty QA Data Synthesis Framework. We design a three-stage pipeline that distills seed entities, synthesizes QA through dual complementary pipelines, and produces a high-quality training set via a downstream optimization and validation module.}
  \label{fig:framework}
\end{figure}

DeepSearch targets long-horizon, multi-hop open-domain retrieval QA: given a
question, the agent must autonomously retrieve, cross-verify, and produce a
traceable answer over many rounds of \texttt{search}/\texttt{browse}.
Conventional multi-hop QA data lacks structural difficulty constraints and
suffers a domain shift from online retrieval, making it ill-suited for training
such agents by RL.

\paragraph{Data Pipeline.}
We synthesize high-difficulty multi-hop QA through two complementary pipelines
(Figure~\ref{fig:framework}). Pipeline~I seeds from obscure Wikipedia entities
and runs breadth-first search over intra-entity hyperlinks to build a directed
acyclic graph (DAG) whose depth sets the number of reasoning hops; questions are
synthesized from \emph{chain} subgraphs (sequential reasoning) and \emph{split}
subgraphs (parallel reasoning and cross-verification), with entities generalized
and answers anchored to fine-grained objective attributes. Pipeline~II uses a
ReAct agent to explore the web in real time, linking cross-domain entities that
share a key attribute and synthesizing from a dual-constraint chain joined by an
attribute-coincidence edge---so that only the intersection of the two chains
determines the answer. Both pipelines pass through a unified post-processing
module that rewrites directly-searchable clues and validates correctness,
uniqueness, and difficulty.

\paragraph{Training Pipeline.}
We train DeepSearch with \textbf{online reinforcement learning}: the agent rolls
out \texttt{search}/\texttt{browse} trajectories in a live retrieval environment,
explores through trial and error, and updates its policy on the fly from the
correctness of its final answer. We use
Group Relative Policy Optimization (GRPO) with an outcome-based reward model
(ORM), following a curriculum from moderate- to high-difficulty samples.
For context management, rather than naively truncating or clearing the
history~\cite{deepseekv32}, we implement it through four mechanisms:
\emph{sliding-window observation filtering} (keeping only the most recent
observation rounds while retaining the full reasoning trajectory), \emph{context
restart via progress summary} (compressing the trajectory into a structured
summary before discarding, so a restart continues from existing progress), a
\emph{dynamic adaptive threshold} (raising the restart point as retrieval
proceeds), and \emph{answer self-check} (reviewing the answer against all
constraints and redoing the task with feedback if any are unmet).

\subsection{Code Agent RL}
To enhance the coding capabilities of system-level agents, we establish an end-to-end coding-agent post-training pipeline. Centered on repository-level software engineering tasks, the pipeline constructs verifiable execution environments, synthesizes high-quality SWE data, and further enhances model capabilities through supervised fine-tuning and agentic reinforcement learning. Our model achieves strong performance on SWE-bench Verified, surpassing models of comparable scale.

\paragraph{Code Agent Infrastructure.} A robust coding infrastructure is the cornerstone of code-agent post-training, underpinning every stage of the pipeline from large-scale trajectory generation and reinforcement learning to reproducible evaluation. To this end, we build an automated lifecycle management system for containerized execution environments, covering environment provisioning, health monitoring, and automatic retry-based recovery. This infrastructure ensures stable state persistence and seamless multi-turn interactions throughout long-horizon coding tasks. By orchestrating these execution environments at scale, the system supports thousands of concurrent task instances without compromising execution stability or reproducibility. Beyond execution scalability, we expose the model to coding workflows instantiated through multiple scaffolds, such as OpenHands~\cite{wang2025openhands} and SWE-Agent~\cite{yang2024sweagentagentcomputerinterfacesenable}. Such cross-framework trajectory construction allows the model to learn scaffold-invariant problem-solving behaviors rather than overfitting to formats of specific frameworks, thereby improving robustness across diverse execution environments.

\begin{figure}[htbp]
    \centering
    \includegraphics[width=\linewidth]{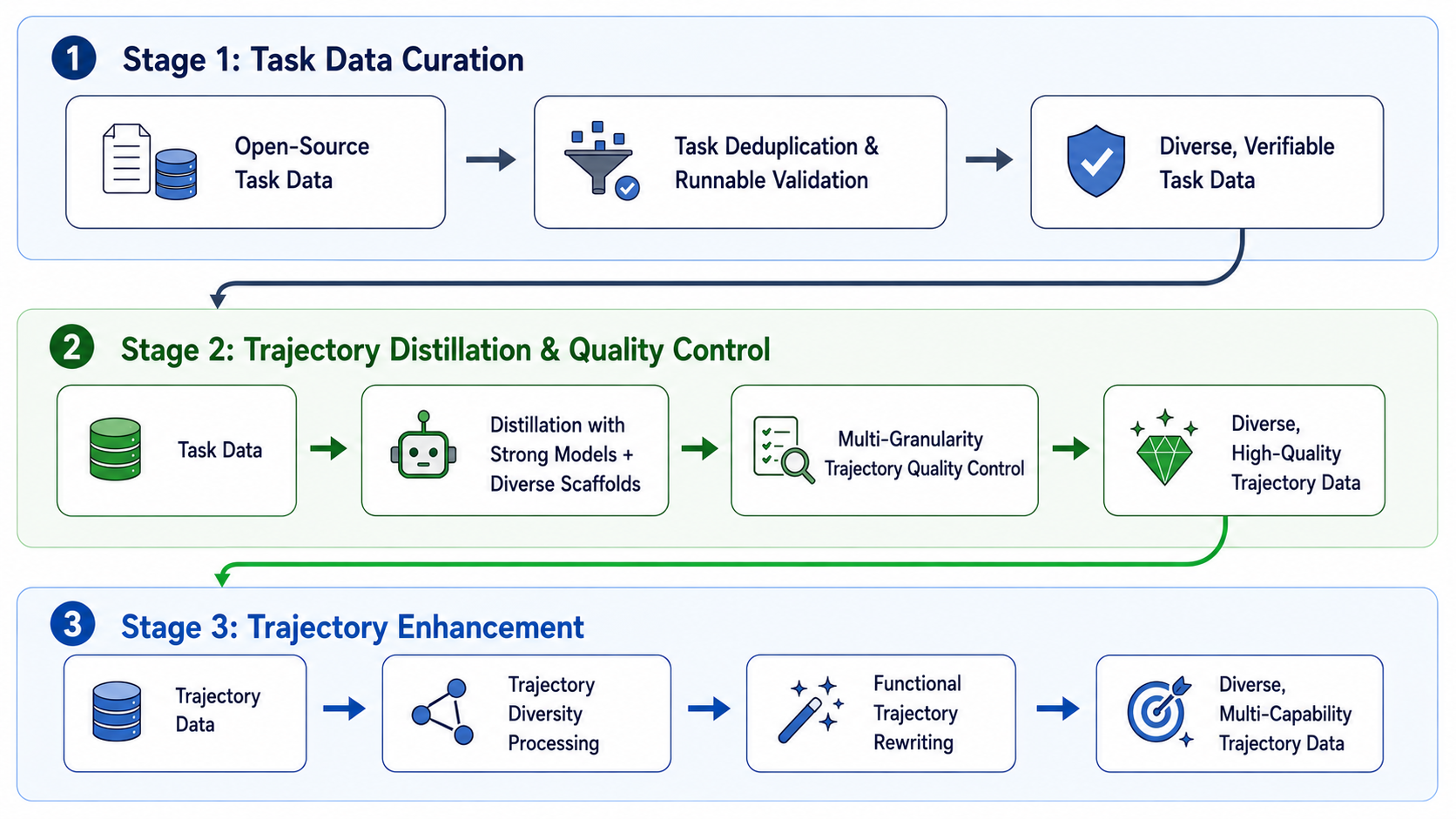}
    \caption{Overview of the Code Agent Data Curation Pipeline. \textbf{Top}: Repository-level SWE tasks are curated into executable and verifiable task instances through deduplication and environment validation. \textbf{Middle}: High-quality trajectories are distilled using strong LLMs across diverse agent scaffolds and filtered via multi-granularity quality control. \textbf{Bottom}: Selected trajectories are further enhanced through diversity augmentation and functional rewriting, producing capability-rich data for code-agent post-training.}
    \label{figure:SWE-data-pipeline}
\end{figure}

\paragraph{Data Curation Pipeline.} Building upon the scalable execution infrastructure, we construct an end-to-end data curation pipeline shown in Figure~\ref{figure:SWE-data-pipeline}. For SWE tasks, we collect a large corpus of repository-level instances with executable environments \cite{zhao2026immersiongithubuniversescaling, jain2025r2e-gym, yang2025swesmith}. Only executable and verifiable environments are retained through fail-to-pass validation.

Based on curated instances, we leverage multiple code-focused LLMs to automatically generate trajectories across diverse scaffolds. Each rollout is configured with up to 300 turns and a 256K context budget. Only trajectories that successfully resolve the target issue, pass all task-specific test cases, and satisfy strict trajectory-level quality criteria with respect to turn/context constraints and tool-call format validity are retained for SFT.

After trajectory generation, we further enhance a subset of trajectories. We diversify trajectory tool-call schemas while preserving the original chain-of-thought (CoT), tool calls, and observations, thereby improving robustness across diverse agent frameworks. Furthermore, we perform structured trajectory refinement by rewriting the initial step with explicit task-level planning and refining selected subsequent CoT segments. This enhancement strengthens planning capabilities and encourages more structured and deliberate reasoning throughout long-horizon SWE workflows. Additionally, we apply the following data-centric post-processing strategies to further improve trajectory quality and training effectiveness.

\begin{itemize}
\item \textbf{Reasoning with Agent-Friendly Template}. We recognize reasoning as a fundamental capability underpinning effective agentic problem solving. To effectively integrate reasoning with agentic tool use in coding tasks, we follow the ReAct paradigm and preserve the full reasoning history at every interaction turn. Meanwhile, JSON serialization introduces excessive character escaping and increases formatting fragility, whereas XML naturally supports flat string representations with substantially lower syntactic overhead \cite{5team2025glm45agenticreasoningcoding}. We therefore adopt an XML-based tool-use template instead of the conventional JSON-style format. We ensure that all tool invocations throughout the trajectories are represented in XML format, thereby mitigating the format mismatch between training and inference.

\item \textbf{Error Masking}. We retain but mask erroneous steps in trajectories to prevent the model from learning incorrect actions, following the widely adopted error-masking strategy \cite{glm5team2026glm5vibecodingagentic, huang2026step35flash, swelego}. Building upon LLM annotations, we further develop a scalable rule-based masking pipeline. A large collection of trajectories is annotated by LLMs to identify representative action-level failure patterns. We then derive error-masking rules calibrated against LLM-generated annotations, enabling erroneous actions to be masked while preserving legitimate diagnostic behaviors, such as test execution and issue reproduction, which may produce abnormal observations but remain essential to the workflow. Empirically, error-masked trajectories yield nearly a 3\% improvement over unmasked data, while our rule-based masking pipeline achieves performance comparable to that of LLM-based annotation with substantially higher efficiency and scalability.
\end{itemize}

\paragraph{Agentic Reinforcement Learning.}
To further enhance the coding agent's capabilities in long-horizon tasks, we introduce agentic reinforcement learning on top of supervised fine-tuning.
Stable and effective reinforcement learning relies on scalable training infrastructure, verifiable task environments, and high-quality data with appropriate difficulty. To this end, we integrate our internal Agent Harness and execution environments into Slime \cite{slime_github}, building a fully asynchronous and decoupled reinforcement learning infrastructure. This design enables environment interaction, trajectory sampling, and policy updates to run in parallel, thereby supporting stable training on large-scale long-horizon coding tasks.
For training data, we first perform difficulty-based task filtering according to the current model's pass@8 performance, excluding overly simple instances with a pass@8 rate above 0.9 and overly difficult instances with a pass@8 rate below 0.1. For the reward design, we adopt a standard binary verifiable task-completion reward. For optimization, we build upon GRPO \cite{shao2024deepseekmathpushinglimitsmathematical} and further incorporate training strategies such as TIS \cite{yao2025offpolicy} and Dynamic Sampling \cite{yu2025dapoopensourcellmreinforcement} to improve sampling efficiency and training stability in long-horizon interaction scenarios. In addition, we remove standard-deviation normalization from advantage computation and adopt prompt-level loss aggregation \cite{liu2025understandingr1zeroliketrainingcritical, yu2025dapoopensourcellmreinforcement, khatri2025artscalingreinforcementlearning}. With the above infrastructure, data-filtering mechanism, verifiable reward design, and optimization strategies, reinforcement learning consistently delivers stable performance gains for coding agents.

\subsection{Claw Agent RL}

Claw Agent trains an autonomous agent expert for complex multi-step tasks in isolated sandbox environments. Unlike the tool-use agent setting, which focuses on structured API-style function invocation, Claw Agent targets \emph{environment-interactive} tasks: the model must reason over multi-turn observations, invoke tools in context, handle execution feedback, and complete tasks whose success is determined by verifiable sandbox outcomes. This setting emphasizes long-horizon decision making and end-to-end task completion rather than single-turn tool usage accuracy.

\begin{figure}[htbp]
    \centering
    \includegraphics[width=\linewidth]{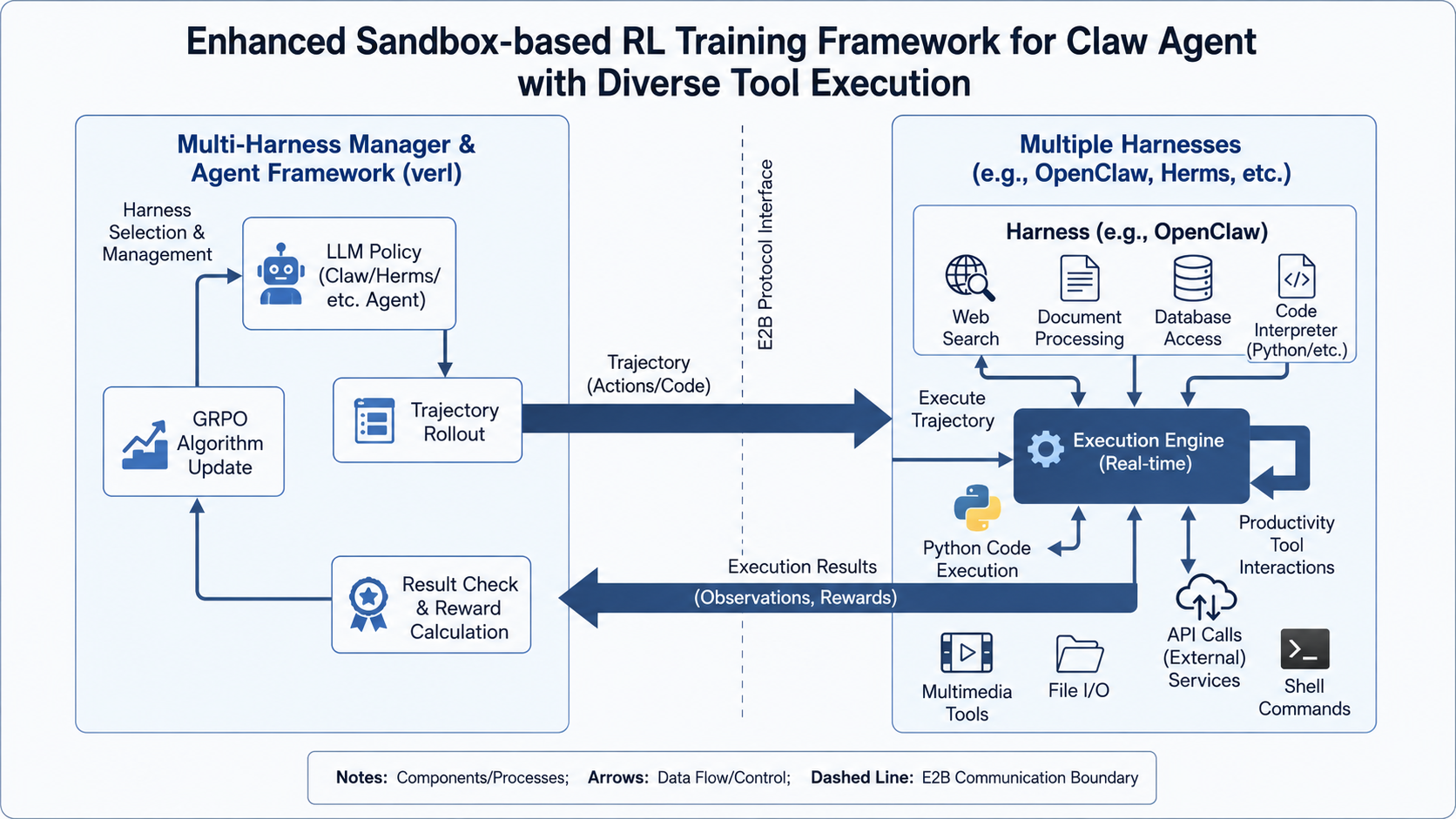}
    \caption{Claw Agent training framework with sandboxed tool execution. 
A multi-harness training loop samples agent trajectories, dispatches actions/code to heterogeneous sandbox harnesses through the E2B protocol interface, and receives execution observations and verification rewards for GRPO-based policy optimization. 
The execution side supports diverse tools including web search, document processing, databases, code interpreters, file I/O, shell commands, multimedia tools, and external APIs.}
\end{figure}

We formulate each task as a partially observable multi-turn decision process. A trajectory $\tau$ consists of interleaved model-generated tokens, tool calls, harness responses, and optional final answers, and receives a scalar reward $R(\tau)$ from task verification. Claw Agent is organized around a stateful agent loop with transitions
\textsc{Pending} $\rightarrow$ \textsc{Generating} $\rightarrow$ \textsc{Processing\_Tools} $\rightarrow$ \textsc{Terminated},
supporting dozens of assistant turns per trajectory as well as parallel execution of multiple tool calls within a single turn. To preserve clean optimization signals, Claw Agent applies token-level response masking so that only model-generated tokens contribute to the policy gradient, while harness-returned content is excluded.

\paragraph{Data Pipeline.} Training data is constructed from a curated collection of real-world analytical tasks paired with sandbox-based verification environments. To improve both coverage and difficulty diversity, we combine tasks from multiple categories spanning a range of complexity levels. The resulting dataset contains two complementary splits: a difficulty-mixed split emphasizing more challenging analytical tasks, and an easy-case split used for curriculum-style bootstrapping.

Each sample provides a structured task specification including:
(1) a system prompt and a user prompt,
(2) task-specific resource provisioning for required dependencies,
(3) a per-sample harness schema injected at inference time,
(4) a \texttt{check\_list} specifying verification criteria and reward weights, and
(5) ground-truth references.
The per-sample harness schema enables fine-grained control over the capabilities exposed to the agent for different task types, improving flexibility and operational safety. It also simplifies the integration of heterogeneous tasks into a unified training pipeline, since tool availability, verification logic, and resource access can all be configured at the sample level.

\paragraph{Claw Agent Reinforcement Learning.} Claw Agent is trained with GRPO using group-wise advantage normalization. For each task prompt, we sample a group of independent trajectories and compute the advantage for trajectory $i$ as:
\begin{equation}
    A_i = \frac{R_i - \mu_g}{\sigma_g + \varepsilon},
\end{equation}
where $R_i$ denotes the task-completion reward obtained via asynchronous verification, and $\mu_g$ and $\sigma_g$ are the mean and standard deviation of rewards within the sampled group. This removes the need for a learned value critic while providing a stable relative baseline for policy optimization. A KL penalty with respect to a reference policy is additionally incorporated to limit excessive policy drift and stabilize training.

Training is performed on Claw expert using a distributed setup spanning multiple nodes with tensor, pipeline, and expert parallelism. The framework supports very long contexts to accommodate long-horizon agent trajectories, while rollout generation is executed asynchronously to improve GPU utilization and reduce stalls caused by environment-side latency.

To further improve credit assignment in long-horizon interactions, Claw Agent introduces a token-level credit assignment mechanism that modulates gradient contributions within each trajectory. Concretely, token-level policy updates are reweighted according to their structural role in the interaction sequence, such as action generation, tool selection, intermediate reasoning, and final answer synthesis. This allows optimization to place greater emphasis on outputs more strongly associated with successful task completion, rather than treating all generated tokens as equally informative.

\subsection{Multi-Teacher On-Policy Distillation (MOPD)}
\label{sec:opd-mopd}

Post-training, and reinforcement learning in particular, has become
the primary lever for advancing a base model's intelligence and
per-domain capability. Fusing multiple such capabilities into a
single generalist, however, remains a hard problem. Reinforcement
learning on a heterogeneous reward mixture, whether staged or jointly
mixed, is prone to a pronounced \emph{see-saw} effect, where gains
on one capability are routinely offset by regressions on others.
It is also expensive: sequencing the specialists imposes a strict
dependency chain, and jointly mixing them enlarges the reward stack
that must be served at every optimization step. To address both
issues, we adopt Multi-Teacher On-Policy Distillation (MOPD): each
training sample is routed to the domain specialist that has already
mastered it, and the student is supervised directly on its own
rollouts through a token-level reverse-KL objective. Relative to
mixed-reward RL, this replaces a heterogeneous reward stack with a
homogeneous, dense distillation signal, allowing new domains to be
folded into the student without eroding those already fused.

\paragraph{Objective.}
Let $\pi_\theta$ be the student and $\pi_{T_k}$ the frozen specialist
for domain $k$. Each training sample carries a routing key
\texttt{teacher\_route} that deterministically selects its supervising
teacher. MOPD minimizes a routed mixture of token-level reverse-KL
distillation losses computed on the student's own rollouts:
\begin{equation}
\mathcal{L}_{\text{MOPD}}(\theta)\;=\;
\mathbb{E}_{(x,k)\sim\mathcal{D}}
\,\mathbb{E}_{y\sim\pi_\theta(\cdot\mid x)}
\frac{1}{|y|}\sum_{t=1}^{|y|}
D_{\mathrm{KL}}\!\big(\pi_\theta(\cdot\mid x,y_{<t})
\,\big\|\,\pi_{T_k}(\cdot\mid x,y_{<t})\big),
\label{eq:mopd}
\end{equation}
To make Eq.~\eqref{eq:mopd} tractable, we
estimate the reverse KL with a single-sample $k_1$ estimator and
optimize a clipped policy-gradient surrogate on top of it that
corrects for the asynchronous rollout-to-training drift induced by our
vLLM inference engine; the full derivation and the magnitude-preserving
monitoring quantity $\mathcal{L}_{\mathrm{abs}}$ we log alongside are
deferred to Appendix~\ref{app:opd-derivation}. Gradients flow only
through the student $\pi_\theta$; the routed teacher $\pi_{T_k}$ is
frozen and provides the soft target.

\begin{figure*}[t]
\centering
\begin{minipage}[t]{0.46\textwidth}
    \vspace{0pt}
    \centering
    \footnotesize
    \setlength{\tabcolsep}{3pt}
    \captionof{table}{MOPD training hyper-parameters used in the
    production run. Distillation-specific settings correspond to the
    estimator and surrogate defined in
    Appendix~\ref{app:opd-derivation}.}
    \label{tab:mopd-hparams}
    \begin{tabular}{@{}llr@{}}
    \toprule
    Group & Parameter & Value \\
    \midrule
    \multirow{4}{*}{Data \& rollout}
        & \texttt{max\_prompt\_length}   & $40{,}960$ \\
        & \texttt{max\_response\_length} & $8{,}192$ \\
        & \texttt{rollout\_n}            & $1$ \\
        & \texttt{shuffle}               & \texttt{True} \\
    \midrule
    \multirow{3}{*}{Optimisation}
        & \texttt{learning\_rate}        & $1.0\times10^{-6}$ \\
        & \texttt{batch\_size}           & $64$ \\
        & \texttt{ppo\_mini\_batch\_size}& $32$ \\
    \midrule
    \multirow{2}{*}{Distillation}
        & \texttt{loss\_mode}            & \texttt{k1} \\
        & \texttt{use\_policy\_gradient} & \texttt{True} \\
    \bottomrule
    \end{tabular}
\end{minipage}%
\hfill
\begin{minipage}[t]{0.5\textwidth}
    \vspace{0pt}
    \centering
    \includegraphics[width=\linewidth]{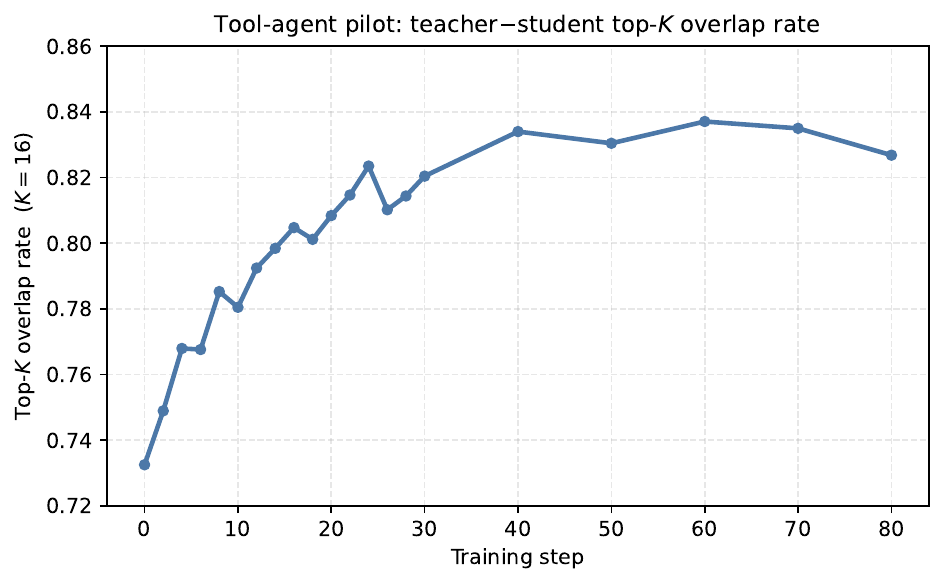}
    \captionof{figure}{Pilot study on the tool-agent domain: the
    teacher\textendash student top-$K$ token overlap rate ($K{=}16$)
    climbs monotonically from $0.73$ to $0.84$ over training.}
    \label{fig:overlap-curve}
\end{minipage}
\end{figure*}

\paragraph{Training configuration.}
The production MOPD run fuses more than ten specialists spanning three categories: \textbf{Reasoning RL}, \textbf{General RL}, and \textbf{Agent RL}. Domains are mixed in a strict $1{:}1$ ratio so that no single capability dominates the gradient signal. The key training hyper-parameters are collected in Table~\ref{tab:mopd-hparams}, with one detail worth highlighting: following the Early Stopping Rollout strategy of \citet{ziheng2026less}, we deliberately cap \texttt{max\_response\_length} at $8$K tokens even for long-form math, code, and deep-search. Empirically this shortens each rollout step, lowers vLLM KV-cache pressure, and cuts total GPU hours substantially, while matching or exceeding runs with much longer response budgets on several domains.

\begin{figure}[t]
    \centering
    \begin{minipage}[t]{0.48\linewidth}
        \centering
        \includegraphics[width=\linewidth]{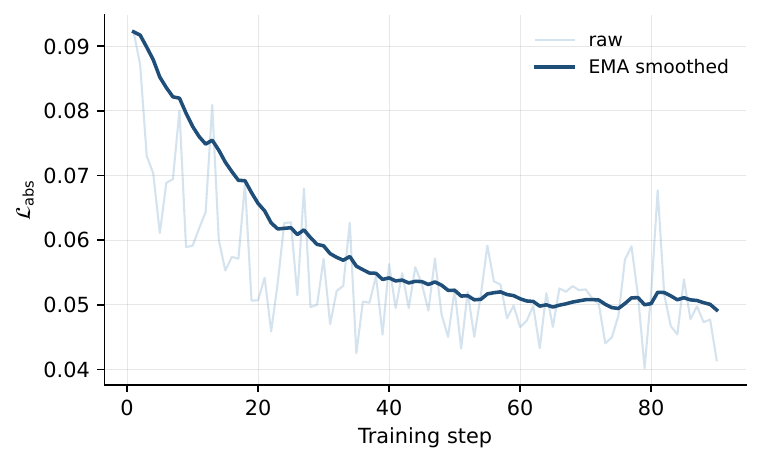}\\[-1pt]
        {\small (a) Monitoring quantity $\mathcal{L}_{\mathrm{abs}}$.}
    \end{minipage}
    \hfill
    \begin{minipage}[t]{0.48\linewidth}
        \centering
        \includegraphics[width=\linewidth]{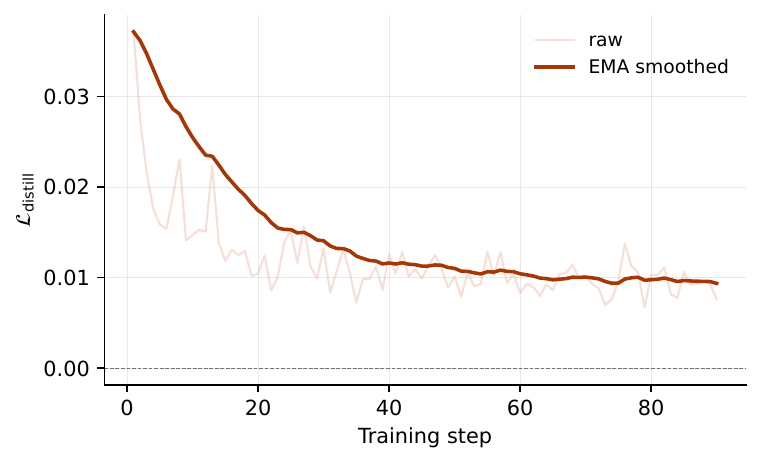}\\[-1pt]
        {\small (b) Back-propagated quantity $\mathcal{L}_{\mathrm{distill}}$.}
    \end{minipage}
    \caption{Distillation training dynamics on the MOPD
             production run. Light traces are per-step values; dark
             traces are exponential moving averages ($\alpha{=}0.9$).
             $\mathcal{L}_{\mathrm{abs}}$ is the magnitude-preserving
             diagnostic (not back-propagated), while
             $\mathcal{L}_{\mathrm{distill}}$ is the clipped
             policy-gradient surrogate that is actually optimized.
             Both decrease monotonically and settle by roughly
             step~$60$. Formal definitions are in
             Appendix~\ref{app:opd-derivation}.}
    \label{fig:distill-curves}
\end{figure}

\paragraph{Convergence and results.}
Throughout the run we monitor two signals: (i) the teacher\textendash
student top-$K$ token \emph{overlap rate}, a direct proxy for how
aligned the student's decoding has become with the teacher, and
(ii) the pair of loss quantities $\mathcal{L}_{\mathrm{abs}}$
(magnitude-preserving, not back-propagated) and
$\mathcal{L}_{\mathrm{distill}}$ (the optimized clipped surrogate)
defined in Appendix~\ref{app:opd-derivation}.
Figure~\ref{fig:overlap-curve} shows a representative pilot on the
tool-agent domain: the top-$K$ overlap rate climbs monotonically from
$0.73$ at initialisation to $0.84$, evidencing that the student's
decoding progressively aligns with the teacher.
Figure~\ref{fig:distill-curves} then shows the two loss curves on the
multi-task production run: both decrease monotonically without
divergent spikes and settle at
$\mathcal{L}_{\mathrm{abs}}\!\approx\!0.05$ and
$\mathcal{L}_{\mathrm{distill}}\!\approx\!0.01$ within roughly $60$
optimization steps, indicating stable convergence under the routed
multi-teacher signal. Two design choices adopted from single-teacher
pilots carry into this production run: (a) matching the teacher's
parameter count to the student yields a markedly higher overlap rate
than a substantially larger teacher, at no measurable loss in final
quality  -- an observation also reported by \citet{li2026rethinking}; and (b) training on prompts the teacher itself had seen
further raises overlap and improves the final student. The detailed
numerical impact of MOPD on every targeted capability is reported in
Section~\ref{sec:evaluation}.

\subsection{Hybrid Median-length Policy Optimization (HMPO)}
\label{sec:token-efficiency}

\begin{figure*}[ht]
\centering
\includegraphics[width=\linewidth]{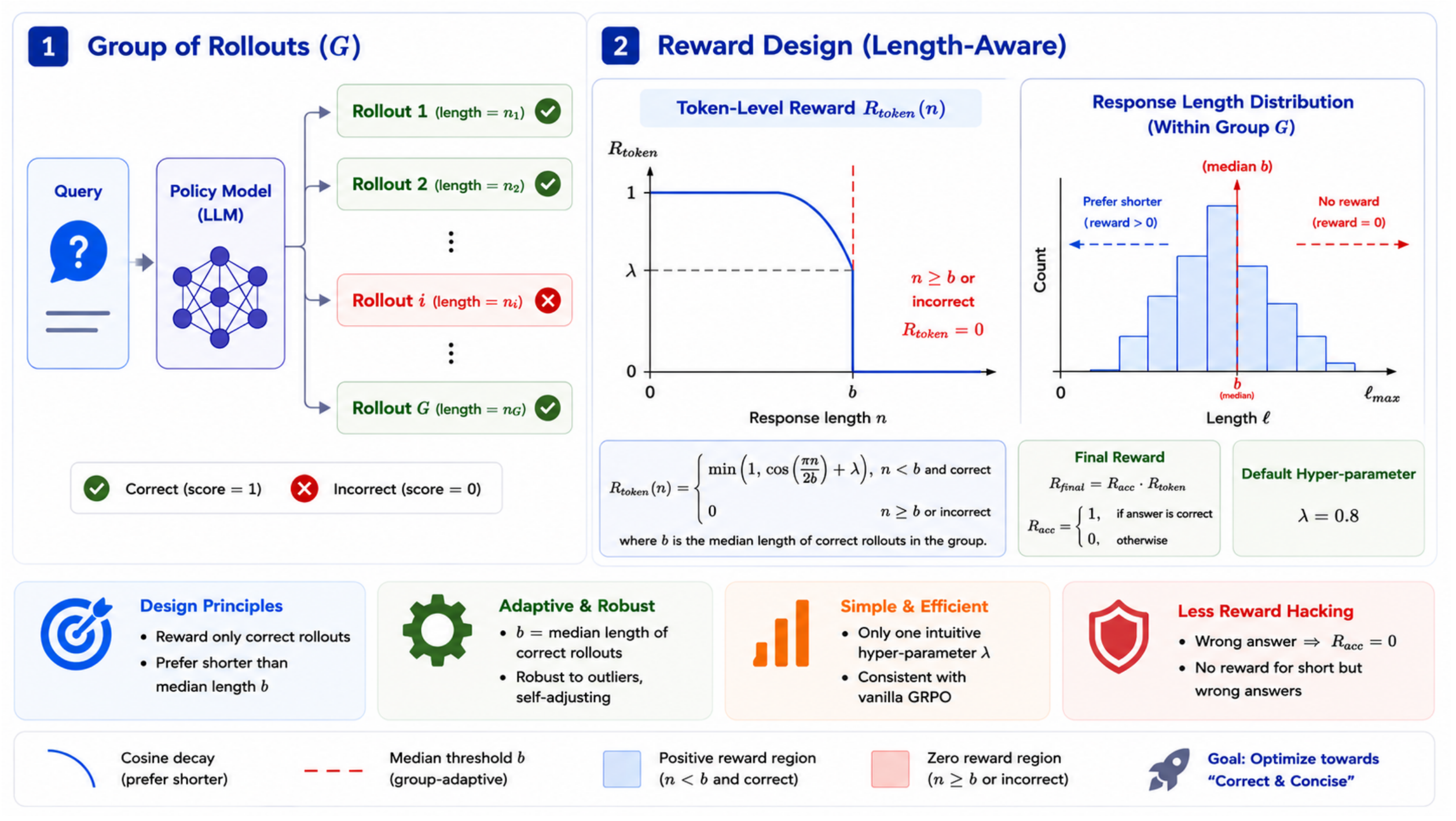}
\caption{Overview of HMPO. \textbf{Left:} For each query, the policy samples a group of rollouts ($G$). \textbf{Right:} Instead of relying on a static threshold, HMPO dynamically derives an adaptive budget $b$ from the median length of only the \emph{correct} rollouts to construct a smooth cosine-decay token reward. \textbf{Bottom:} The final reward is combined multiplicatively to enforce a strict ``correctness-first, length-second'' objective, mathematically preventing reward hacking (i.e., short but incorrect answers strictly receive zero reward).}
\label{fig:method}
\end{figure*}

\paragraph{Why token efficiency.}
The MOPD stage (Section~\ref{sec:opd-mopd}) fuses all per-domain specialists into a single generalist, but it inherits a well-known side effect of strong CoT reasoning models~\cite{wei2022chain,jaech2024openai,guo2025deepseek}: \emph{overthinking}~\cite{chen2025not,sui2025stop}, in which the fused model produces reasoning chains far longer than a task requires, inflating inference latency and serving cost without commensurate accuracy gains. We therefore add a final \emph{token-efficiency} stage that takes the MOPD checkpoint as input and produces the deployed model as output: its sole objective is to compress generation length while preserving, rather than trading away, accuracy.

\paragraph{Method: HMPO.}
To this end we apply \textbf{HMPO} (Hybrid Median-length Policy
Optimization)~\cite{zheng2026hmpo}, a cost-effective \emph{single-stage} RL
framework that, unlike prior length-control methods relying on rigid manual
budgets or expensive multi-stage pruning, dynamically derives a group-adaptive
budget $b$ from the median length of \emph{correct} rollouts and applies a
length-aware token reward. As illustrated in Figure~\ref{fig:method}, HMPO
builds on GRPO and its only change is to redesign the rollout reward $R_i$ to
be budget-aware, through three components.

\emph{(1) Adaptive median budget.}
Static budgets ignore difficulty and prior-rollout estimation is costly, so we
set the budget to the median length of the \emph{correct} rollouts in the
group, $b=\mathrm{median}(\{n_i\mid i\in\mathcal{C}\})$. This isolates the
length signal from failed traces, is difficulty-aware (a harder query yields
longer correct rollouts and thus a looser $b$), and self-tightens as the
policy improves---an implicit curriculum with zero tuning.

\emph{(2) Cosine-decay token reward.}
A naive penalty invites reward hacking, so we reward only correct traces
shorter than $b$ with a smooth landing,
\begin{equation}
R_{\text{token}}=
\begin{cases}
\min\!\left(1,\; \cos\!\left(\frac{\pi n}{2b}\right)+\lambda\right), & \text{if correct and } n<b,\\
0, & \text{otherwise.}
\end{cases}
\end{equation}

\emph{(3) Multiplicative composition.}
$R_{\text{final}}=R_{\text{acc}}\cdot R_{\text{token}}$ enforces a strict
\emph{correctness-first, length-second} hierarchy: incorrect or over-budget
traces receive exactly zero reward, denying efficiency gradients to wrong
answers, whereas additive combination would still pay verbose-but-correct
trajectories. By avoiding any auxiliary stage, the single-pass run uses
$1.5\times$--$2.5\times$ fewer GPU-hours than multi-stage baselines.

\paragraph{Training configuration.}
We run HMPO on a compact subset of roughly 6.5K high-quality mathematics
problems, deliberately mixing samples of varying difficulty and reasoning-chain length so that the group-adaptive budget is exposed to a broad spectrum of correct-rollout lengths. using a group size of $G{=}10$ rollouts per query and a reward offset of
$\lambda{=}0.8$. Notably, although training is conducted \emph{only} on
mathematics, the learned length-control behavior generalizes to unseen
domains---code generation, science QA, and instruction following---indicating
that HMPO instills a general ``answer as concisely as correctness allows''
policy rather than a math-specific shortcut.

\section{Experimental Results}
\label{sec:evaluation}
\subsection{Overall Results}
\begin{table}[t]
  \centering
  \caption{Main evaluation across capability axes. Scores marked with $^{*}$ are
  taken from the corresponding official technical reports; all other scores are
  reproduced by our own evaluation harness under
  identical settings. Official numbers are used when available; our reproduced
  numbers are used otherwise. Best per row in \textbf{bold}.}
  \label{tab:main-eval}
  \scriptsize
  \setlength{\tabcolsep}{2.5pt}
  \resizebox{\linewidth}{!}{%
  \begin{tabular}{@{}l cccccccc@{}}
    \toprule
    & \shortstack{\textbf{Mach-Mind-4}\\\textbf{Flash}}
    & \shortstack{GLM-4.7-Flash\\30B-A3B}
    & \shortstack{Qwen3.5\\35B-A3B}
    & \shortstack{Nemotron-3\\120B-A12B}
    & \shortstack{Qwen3.5\\122B-A10B}
    & \shortstack{Step-3.5-Flash\\196B-A11B}
    & \shortstack{MiMo-V2-Flash\\309B-A15B}
    & \shortstack{Kimi-K2.5\\1T-A32B} \\
    \midrule
    \multicolumn{8}{@{}l}{\textit{Reasoning \& Knowledge}}\\
    AIME'26            & 92.70 & 88.65 & 91.87 & 89.90 & 91.67 & \textbf{94.17} & 93.75 & 93.30 \\
    AIME'25            & 92.08 & 91.60* & 88.75 & 90.83 & 89.48 & \textbf{97.30*} & 94.10* & 96.10* \\
    LiveCodeBench-V6   & 80.91 & 64.00* & 74.60* & 78.44* & 78.90* & \textbf{86.40*} &  80.60* & 85.00* \\
    GPQA-Diamond       & 83.08 & 75.20* & 84.20* & 79.42* & 86.60* & 83.33 & 83.70* & 87.60* \\
    \midrule
    \multicolumn{8}{@{}l}{\textit{Instruction Following \& Writing}}\\
    IFEval             & \textbf{94.64} & 85.86 & 91.90* & 92.14 & 93.40* & 94.09 & 88.91 & 94.32 \\
    IFBench            & \textbf{82.82} & 56.89 & 70.20* & 73.30* & 76.10* & 65.22 & 60.88 & 67.43 \\
    LexInstructEval    & \textbf{74.63} & 42.79 & 71.56 & 63.27 & 74.38 & 65.62 & 59.43 & 64.69 \\
    WritingBench       & 74.61 & 70.92 & 71.61 & 77.10 & 74.80 & 80.34 & 75.24 & \textbf{81.47} \\
    \midrule
    \multicolumn{8}{@{}l}{\textit{Safety}}\\
    Content-SafetyBench    & \textbf{98.20} & 95.38 & 97.00 & 95.40 & 95.27 & 94.59 & 92.60 & 95.98 \\
    Behavioral-SafetyBench & \textbf{80.74} & 29.55 & 28.23 & 23.28 & 29.92 & 24.32 & 35.88 & 67.75 \\
    \midrule
    \multicolumn{8}{@{}l}{\textit{Tool-Use \& Code Agent}}\\
    BFCL-v4            & 75.80 & 68.40 & 67.30* & 55.80 & 72.20* & 72.30 & \textbf{76.30*} & 74.50  \\
    $\tau^2$-bench     & 80.04 & 79.50* & 81.20* & 60.46* & 79.50* & \textbf{88.20*} & 80.93 & 72.82 \\
    SWE-bench Verified & 70.60 & 59.20* & 69.20* & 60.50* & 72.00* & 74.40* & \textbf{80.20*} & 76.80* \\
    \midrule
    \multicolumn{8}{@{}l}{\textit{DeepSearch}}\\
    BrowseComp         & 61.70 & 42.80* & 61.00* & 31.28* & \textbf{63.80*} & 51.60* & 45.40* & 60.60* \\
    BrowseComp-zh      & \textbf{72.31} & 54.67 & 69.50* & 31.60* & 70.30* & 66.90* & 69.50* & 71.28 \\
    X-Bench            & 65.00 & 52.00 & 61.00* & 51.00 & 73.00 & \textbf{83.70*} & 61.00* & 64.00 \\
    \midrule
    \multicolumn{8}{@{}l}{\textit{OpenClaw}}\\
    PinchBench         & 74.90 & 64.60 & 73.70 & 6.80 & \textbf{80.60*} & 76.60* & 65.50 & 79.60* \\
    ClawBench (pass@3)  & 84.20 & 66.20 & 83.40 & 23.50 & \textbf{85.11} & 83.97 & 70.60 & 82.20 \\
    ClawEval           & 69.49 & 58.70 & 61.12 & 59.49 & 68.20 & 68.30 & 61.55 & \textbf{74.90*} \\
    \bottomrule
    \end{tabular}

  }
\end{table}

We evaluate \textbf{Mach-Mind-4-Flash} against a broad set of open- and closed-weight baselines spanning a wide activated-parameter range: trillion-scale MoE models (Kimi-K2.5-1T-A32B~\cite{team2026kimi}), MiMo-V2-Flash-309B-A15B~\cite{xiao2026mimo} $\sim$120B models (Qwen3.5-122B-A10B~\cite{qwen3.5}, Nemotron-3-Super-120B-A12B~\cite{chandiramani2026nemotron}), and compact $\le$35B models (Qwen3.5-35B-A3B~\cite{qwen3.5}, GLM-4.7-Flash-30B-A3B~\cite{5team2025glm45agenticreasoningcoding}). To ensure a self-consistent comparison, we use our own evaluation harness under identical decoding settings for all models; official numbers are adopted when available, and our reproduced values are used otherwise. The benchmark suite covers eight capability axes: reasoning \& knowledge, instruction following, writing, safety, tool-use, code agent ability, DeepSearch and OpenClaw. For Reasoning \& Knowledge, math benchmarks are evaluated with avg@32 and other benchmarks with avg@4. 

\paragraph{Reasoning and Knowledge.}
We probe reasoning and knowledge with competition mathematics (AIME'25, AIME'26)~\cite{aime25,aime26}, contamination-resistant code generation (LiveCodeBench-V6)~\cite{jain2025livecodebench}, and graduate-level science (GPQA-Diamond)~\cite{rein2023gpqa}. The headline result is on competition math: Mach-Mind-4-Flash scores 92.70 on AIME'26 and 92.08 on AIME'25, matching the $\sim$120B Qwen3.5-122B-A10B (91.67) and narrowing the gap to trillion-scale Kimi-K2.5 (93.30) to under one point---strong evidence that aggressive post-training compression does not sacrifice multi-step reasoning depth. On code generation it reaches 80.91 on LiveCodeBench-V6, ahead of all same-scale models and trailing only Step-3.5-Flash and Kimi-K2.5. GPQA-Diamond (83.08) tells a similar story: competitive with 120B-class baselines despite 30$\times$ fewer activated parameters.

\paragraph{Instruction Following and Writing.}
We assess instruction following with IFEval~\cite{zhou2023instruction} (verifiable constraints), IFBench~\cite{pyatkin2026generalizing} (held-out, out-of-domain constraints), and our in-house LexInstructEval~\cite{ren2026lexinstructeval}, which uses a rule-based programmatic engine to verify compliance with compositional lexical constraints without LLM-as-a-judge bias. Open-ended generation is evaluated on WritingBench~\cite{wu2026writingbench}, spanning 6 primary and 100 fine-grained domains with instance-specific multi-dimensional criteria. 
Instruction following is the model's standout strength: Mach-Mind-4-Flash tops all three benchmarks with IFEval 94.64, IFBench 82.82, and LexInstructEval 74.63. The IFBench lead is especially telling---many baselines that score well on IFEval drop sharply on held-out constraints, indicating overfitting to known templates rather than genuine constraint comprehension. On WritingBench, Mach-Mind-4-Flash reaches 74.61, on par with Qwen3.5-122B (74.80) and ahead of most same-scale models. 

\paragraph{Safety.}
We design \textbf{Content-SafetyBench} and \textbf{Behavioral-SafetyBench} for
evaluating LLM safety. Content-SafetyBench is constructed from safety evaluation
data accumulated through extensive internal testing, while Behavioral-SafetyBench
is built from internal evaluation data and further augmented with representative
scenarios derived from Agent-SafetyBench~\cite{zhang2024agent}, providing
complementary coverage of content-level and behavior-level safety risks. Safety
is another clear strength of Mach-Mind-4-Flash: it tops both benchmarks, reaching
98.20 on Content-SafetyBench and 80.74 on Behavioral-SafetyBench. The
Behavioral-SafetyBench result is particularly striking---Mach-Mind-4-Flash leads the
runner-up (Kimi-K2.5, 67.75) by around 13 points, while most baselines fall in the
20--35 range, revealing that behavior-level safety under agentic settings remains
an unsolved challenge for the field and a distinctive advantage of our model.

\paragraph{Tool Use and Code Agent.}
We evaluate tool-use ability with BFCL-v4~\cite{patil2025berkeley} (tool calling) and $\tau^2$-bench~\cite{barres2025tau} (multi-turn decentralized control), and code agent ability with SWE-bench Verified~\cite{yang2025swesmith} (repository-level issue resolution). On BFCL-v4, Mach-Mind-4-Flash scores 75.80, outperforming all same-scale 35B models (Qwen3.5-35B 67.30, GLM-4.7-Flash 68.40) and matching the current SOTA MiMo-V2-Flash (76.30), while surpassing much larger models including Qwen3.5-122B (72.20) and Kimi-K2.5 (74.50). On $\tau^2$-bench it reaches 80.04, substantially ahead of Kimi-K2.5 (72.82) and Qwen3.5-122B (79.50), indicating strong multi-turn coordination ability. On SWE-bench Verified it scores 70.60, comparable to Qwen3.5-122B (72.00) and trailing only models with dedicated SWE optimization pipelines.

\paragraph{Deep Search.}
We assess deep-search ability with BrowseComp / BrowseComp-zh~\cite{wei2025browsecomp,zhou2025browsecomp} and X-Bench~\cite{chen2025xbench} (persistent multi-constraint web browsing). Mach-Mind-4-Flash tops BrowseComp-zh at 72.31, ahead of Qwen3.5-122B (70.30) and Kimi-K2.5 (71.28), and remains competitive on English BrowseComp (61.70) and X-Bench (65.00). This axis shows the sharpest divide across all baselines: weaker models collapse below 45 points, confirming that long-horizon web browsing with constraint tracking remains an unsolved bottleneck for the field.

\paragraph{Claw Agent.}
We assess broader autonomous-agent competence with three OpenClaw benchmarks: ClawBench~\cite{zhang2026clawbench} (everyday online tasks), ClawEval~\cite{ye2026claw} (trustworthy agent evaluation with trajectory-transparent grading), and PinchBench (coding agents in an agentic loop). Mach-Mind-4-Flash scores 84.20 on ClawBench (pass@3) and 69.49 on ClawEval. On ClawBench it outperforms Kimi-K2.5 (82.20) and Step-3.5-Flash (83.97), trailing only Qwen3.5-122B (85.11). On ClawEval it trails Kimi-K2.5 (74.90) but outperforms Qwen3.5-122B (68.20) and all other baselines. On PinchBench it scores 74.90, trailing the larger Kimi-K2.5 (79.60) and Qwen3.5-122B (80.60) but ahead of all same-scale models.

\subsection{Effect of Expert Training and MOPD Fusion}
To validate the effectiveness of our post-training pipeline, we report the performance progression from the SFT base model, through domain-specific expert teachers, to the final MOPD-fused model across our three RL tracks: Reasoning, General, and Agent. Results are summarized in Table~\ref{tab:MOPD-progression}.

\begin{table}[t]
  \centering
  \caption{Performance progression across the post-training pipeline. \textbf{SFT Base}: the shared SFT checkpoint (Qwen3.5-35B-A3B after SFT). \textbf{Expert Teacher}: the domain-specific RL specialist. \textbf{MOPD Final}: the unified model after multi-teacher on-policy distillation. Each expert improves or maintains performance on its target domain; MOPD consolidates these gains into a single model.}
  \label{tab:MOPD-progression}
  \small
  \setlength{\tabcolsep}{4pt}
  \begin{tabular}{@{}llccc@{}}
    \toprule
    \textbf{RL Track} & \textbf{Benchmark} & \textbf{SFT Base} & \textbf{Expert Teacher} & \textbf{MOPD Final} \\
    \midrule
    \multirow{1}{*}{Reasoning}
    & LiveCodeBench-V6   & 79.39 & 80.23 & 80.12 \\
    \midrule
    \multirow{3}{*}{General}
    & IFEval             & 92.42 & 94.64 & 94.84 \\
    & IFBench            & 72.79 & 82.65 & 82.92 \\
    & LexInstructEval    & 72.21 & 75.47 & 75.63 \\
    \midrule
    \multirow{4}{*}{Agent}
    & SWE-bench Verified & 69.00 & 73.80 & 71.10 \\
    & PinchBench         & 74.31 & 77.10 & 75.90 \\
    & ClawBench (pass@3)  & 80.70 & 80.30 & 83.20 \\
    & ClawEval           & 61.82 & 67.23 & 70.35 \\
    \bottomrule
  \end{tabular}
\end{table}

\paragraph{Reasoning.}
The Reasoning RL expert improves LiveCodeBench-V6 from 79.39 (SFT base) to 80.23. After MOPD fusion, the fused model achieves 80.12, on par with the specialist. Notably, ablation experiments show that removing the Reasoning expert from the MOPD teacher pool causes a 2--4\% drop on reasoning benchmarks, confirming that this expert serves as a critical anchor preventing capability regression during multi-domain fusion. Detailed ablations are deferred to Appendix~\ref{app:opd-ablation}.

\paragraph{General.}
The General RL expert delivers substantial gains over the SFT base, lifting IFBench from 72.79 to 82.65 (+9.86) and IFEval from 92.42 to 94.64. After MOPD fusion, these gains are fully retained: the fused model reaches 94.84 on IFEval, 82.92 on IFBench, and 75.63 on LexInstructEval, matching or marginally exceeding the expert on all instruction-following benchmarks and confirming that multi-teacher distillation introduces no dilution on this axis.

\paragraph{Agent.}
The Agent RL experts improve over the SFT base on their respective tasks: the Code Agent expert reaches 73.80 on SWE-bench Verified (+4.80), while the Claw Agent expert achieves 77.10 on PinchBench (+2.79). After MOPD fusion, SWE-bench retains most of the gain at 71.10, with a small gap (~2.7 points) reflecting the challenge of preserving highly specialized long-horizon behaviors during multi-domain distillation. Interestingly, the MOPD-fused model \emph{surpasses} individual Agent experts on ClawBench (83.20 vs.\ 80.30) and ClawEval (70.35 vs.\ 67.23). We attribute this to cross-domain transfer: capabilities from the tool-use, instruction-following, and reasoning experts complement the Claw Agent specialist, providing stronger general reasoning that benefits open-ended autonomous tasks.

\paragraph{Summary.}
Across the three tracks, we observe three distinct fusion outcomes: (1) \emph{capability anchoring} for Reasoning, where the expert prevents regression during fusion; (2) \emph{full retention} for General, where the fused model matches the expert; and (3) \emph{mixed results} for Agent, combining partial retention on SWE with positive transfer on OpenClaw. This validates the MOPD design: instead of the see-saw degradation typical of mixed-RL training, on-policy distillation from routed experts consolidates diverse capabilities with minimal interference and occasional synergy.

\begin{figure}[h]
    \centering
    \includegraphics[width=0.8\columnwidth]{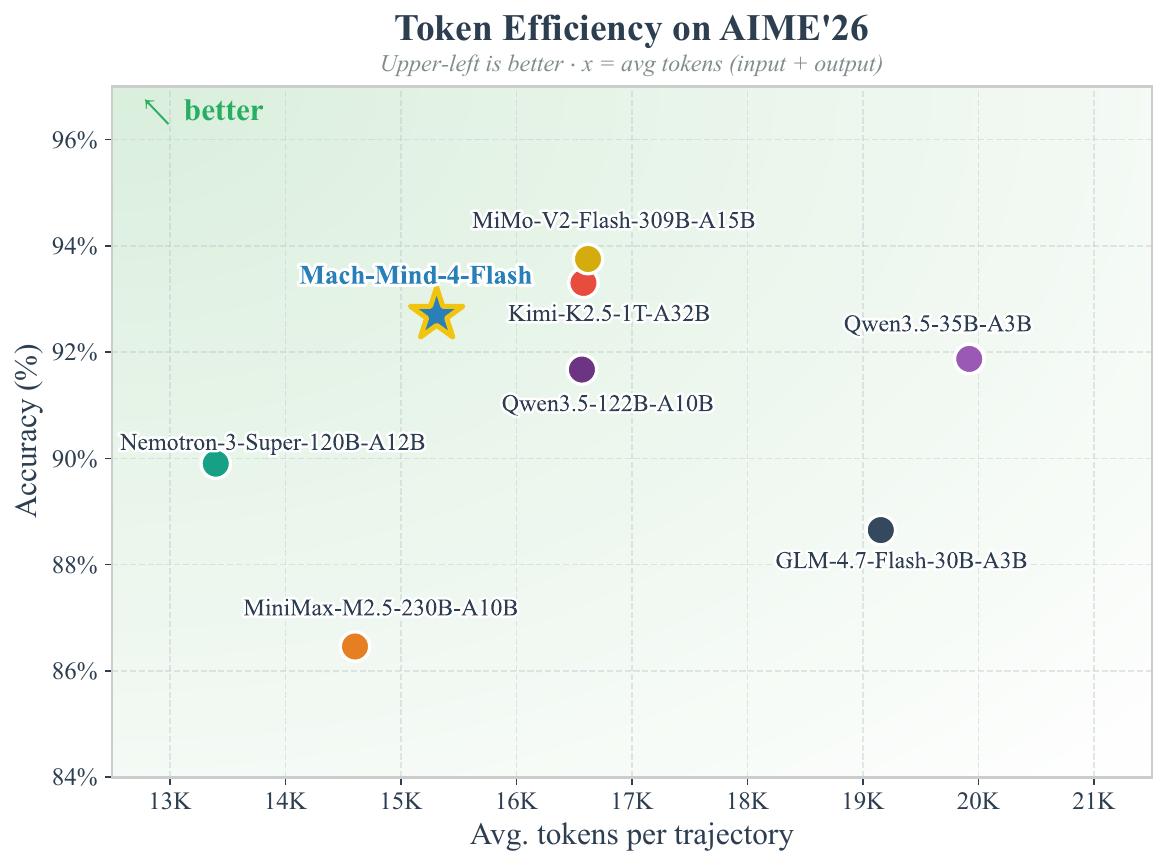} 
    \caption{Token efficiency on AIME'26. Each point represents a model's accuracy vs.\ average tokens per trajectory (input + output). Upper-left is better. HMPO-trained models (stars) achieve competitive or superior accuracy with significantly fewer tokens than frontier models of much larger activated scale.}
    \label{fig:token_efficiency_aime}
\end{figure}

\subsection{Token efficient (HMPO)}
To contextualize HMPO's practical value, Figure~\ref{fig:token_efficiency_aime} plots accuracy versus average tokens per trajectory on AIME'26. Beyond self-improvement, Mach-Mind-4-Flash fundamentally reshapes the Pareto frontier against substantially larger frontier models. It decisively outperforms Nemotron-3-Super-120B-A12B (89.90\% at 13.4K) and delivers performance highly competitive with massive models like Kimi-K2.5-1T-A32B (93.30\% at 16.6K). This demonstrates that HMPO enables reasoning models to punch significantly above their weight class, delivering top-tier mathematical performance at a fraction of the frontier inference cost.

\section{Limitation and Future Work}

In this work, we have presented Mach-Mind-4-Flash, a 35B MoE model that achieves frontier-level performance through a specialization-then-integration post-training pipeline. The capabilities of Mach-Mind-4-Flash mainly come from three sources: the strong initialization provided by Qwen3.5-35B-A3B, a scalable training infrastructure that supports parallel expert development with operator-level acceleration, and Multi-Teacher On-Policy Distillation that consolidates heterogeneous expert capabilities into a single deployable model without the see-saw degradation of mixed-reward RL.

In our effort to push a compact model toward trillion-parameter-level performance, we identified several remaining challenges. First, MOPD fusion introduces a small but consistent gap on extremely long-horizon tasks such as repository-level software engineering, where scaffold-specific behaviors are partially smoothed during distillation. Second, our token-efficiency method (HMPO) currently targets single-turn reasoning; extending budget-aware compression to multi-turn agentic trajectories---where the model must allocate effort across dozens of tool-call turns---remains an open problem. Third, persistent multi-constraint web browsing (DeepSearch) and long-context comprehension remain the weakest axes for compact models, suggesting that scaling the agent horizon alone is insufficient without also improving the model's ability to maintain coherent state over very long interactions. In future work, we will focus on turn-aware token efficiency for agentic settings, improved distillation strategies that better preserve long-horizon expert behaviors, and multimodal extensions that enable visual grounding and GUI interaction within the same post-training framework.

\section{Contributions}

\renewcommand{\thefootnote}{\fnsymbol{footnote}}

\paragraph{Core Contributions.}
Hongxu Chen, Xingru Chen, Baolan Gao, Maokui He, Jiaxin Li, Hao Ma\footnotemark[1], Ziyu Peng, Xiaoyang Qu, Huimin Ren, Ming Song, Tong Sun, Zhaohong Sun, Wenqi Tang, Heqing Wang, Xin Wang, Ze Wang, Qifang Wu, Qiran Xu, Shuling Yang\footnotemark[1], Shengyu Yao, Jiqing Zhan\footnotemark[1], Hu Zhang, Qi Zhang, Rongbin Zhang, Minghui Zheng, Chunpeng Zhou, Xin Zhou\footnotemark[1], Yawei Zhou, Jiaxu Zhu, Xuhan Zhu, Yun Zhu

\paragraph{Contributions.}
Jianhang Chen\footnotemark[2], Jiawei Chen\footnotemark[1], Chaoqun Du, Wuxuan Gong\footnotemark[1], Chenhe Gu, Feng Gu, Jiabang He, Jinkun Hou\footnotemark[2], Miao Huang, Yan Liang\footnotemark[2], Yunze Lin, Chaofan Liu\footnotemark[2], Zhuo Liu\footnotemark[2], Chen Lu, Denghui Lu, Chang Ren, Weidong Shi, Xintian Shen\footnotemark[1], Tianyu Zhang\footnotemark[2], Motong Zhang\footnotemark[2], Yu Zhang, Linhui Yu\footnotemark[2], Yuwen Wang\footnotemark[2], Qian Xiong, Jiale Zhao\footnotemark[2], Linhui Zhou\footnotemark[2]

\paragraph{Project Leaders.}
Hongliang Chen, Zhichao Wang, Lian Wen, Hongsheng Xin, Pengfei Yu, Kaike Zhang\footnotemark[1]

\paragraph{Supervisors.}
Kun Zhan, Bing Zhang, Pan Zhou

\footnotetext[1]{Work done before departure from the team.}
\footnotetext[2]{Intern.}

\bibliographystyle{unsrtnat}

\bibliography{references}

\newpage
\appendix
\section*{Appendix}       
\addcontentsline{toc}{section}{Appendix}   

\section{Acceleration of the Infra operator}
\begin{figure}[h]
    \centering
    \includegraphics[width=0.8\textwidth]{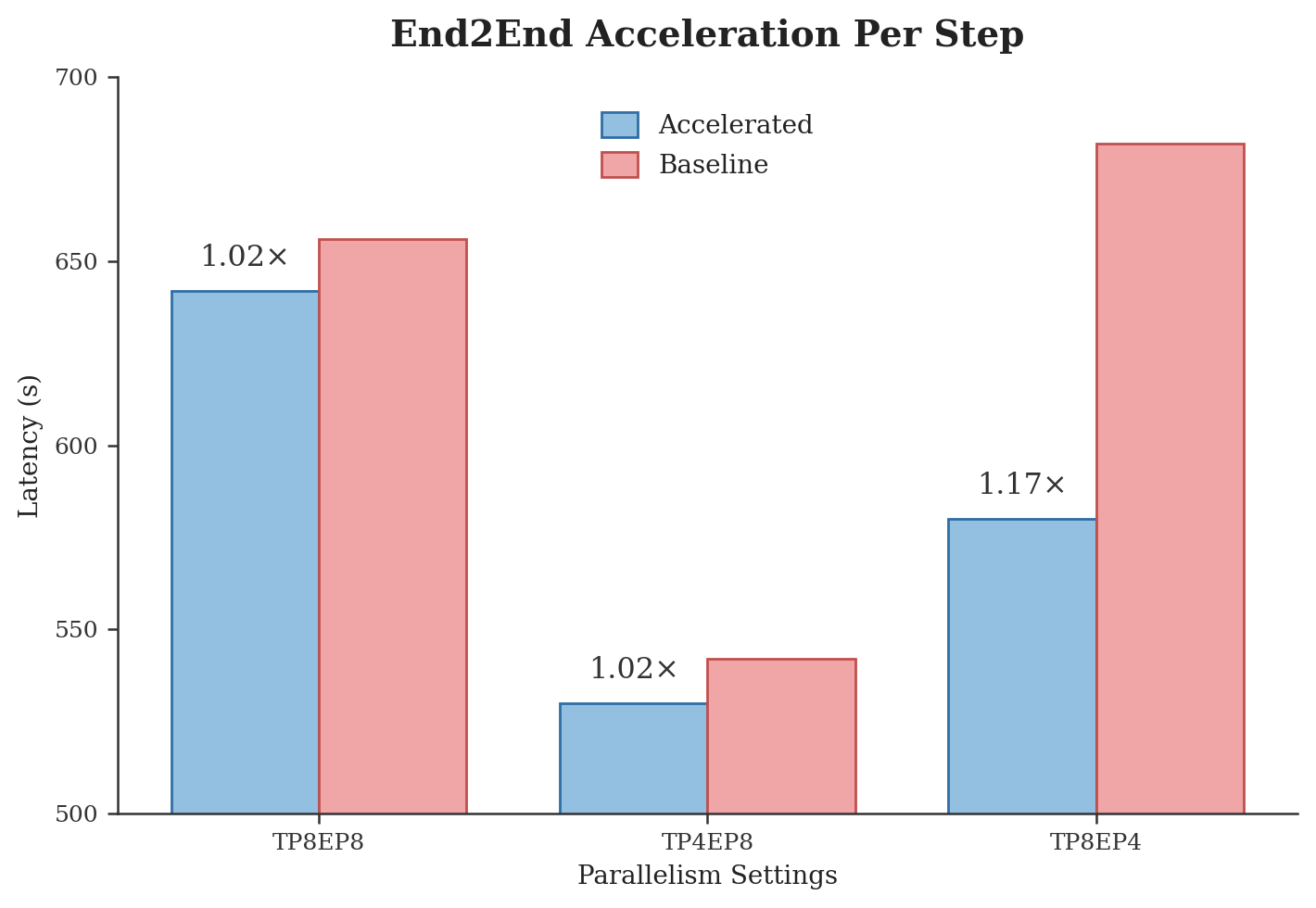} 
    \caption{End-to-end acceleration effect diagram.}
    \label{fig:acceleration_effect}
\end{figure}

Section~\ref{sec:infra-operator-acceleration} describes the fused operator leveraged during training for acceleration, which is designed for the MoE MLP module to reduce the communication and computation overhead. This appendix displays the end-to-end acceleration effects of the operator on the Qwen3.5-35B-A3B model under different parallel parameter configurations. As shown in Figure~\ref{fig:acceleration_effect}, when configured with tp=8, ep=8, the operator accelerates by 2\% (656s/step $\rightarrow$ 642s/step); when configured with tp=4, ep=8, the operator accelerates by 2\% (542s/step $\rightarrow$ 531s/step); when configured with tp=8, ep=4, the operator accelerates by 17\% (681s/step $\rightarrow$ 580s/step). This indicates that the lower the proportion of communication time, the more pronounced the overall acceleration effect of the operator.

\section{Derivation of the MOPD Objective}
\label{app:opd-derivation}

Section~\ref{sec:opd-mopd} (Eq.~\ref{eq:mopd}) writes the MOPD loss as a
routed mixture of token-level reverse KL divergences under the
student's on-policy distribution. This appendix records the three
quantities we defer for brevity: the single-sample estimator we use in
place of the exact reverse KL, the magnitude-preserving diagnostic we
log for monitoring, and the clipped policy-gradient surrogate we
actually back-propagate. Throughout, $\pi_\theta$ denotes the current
student (training engine), $\pi_{T_k}$ the frozen teacher for the
domain of the sample under a routing key
$k=\texttt{teacher\_route}$, and $y\sim\pi_{\theta_{\mathrm{old}}}$ a
rollout of length $|y|$ produced by the (asynchronously updated)
inference engine.  Expectations over $(x,k)\!\sim\!\mathcal{D}$ and
over the length $|y|$ are suppressed to keep the token-level form
prominent.

\paragraph{(a) Single-sample $k_1$ estimator of the reverse KL.}
For a single realised token $y_t \sim \pi_\theta(\cdot\mid x,y_{<t})$,
Schulman's low-variance $k_1$ estimator of the per-token reverse KL is
\begin{equation}
\label{eq:k1}
\widehat{D}^{\,k_1}_{\mathrm{KL}}\!\big(\pi_\theta\,\|\,\pi_{T_k}\big)_{\!t}
\;=\;
\log\pi_\theta(y_t\mid x,y_{<t})
\;-\;
\log\pi_{T_k}(y_t\mid x,y_{<t}),
\end{equation}
which is unbiased for
$D_{\mathrm{KL}}\!\big(\pi_\theta\,\|\,\pi_{T_k}\big)$
when $y_t$ is drawn from $\pi_\theta$. Because Eq.~\eqref{eq:k1}
requires only the two scalars
$\log\pi_\theta(y_t\mid x,y_{<t})$ and
$\log\pi_{T_k}(y_t\mid x,y_{<t})$ per position, the teacher's forward
pass returns a single log-probability per student-sampled token, and
only this per-token scalar is shipped across the
teacher$\rightarrow$student boundary; no full-vocabulary distribution
or top-$K$ list is materialised at loss-computation time.
Substituting Eq.~\eqref{eq:k1} into Eq.~\eqref{eq:mopd} gives the
sampled, tractable form of the MOPD loss.

\paragraph{(b) Magnitude-preserving diagnostic $\mathcal{L}_{\mathrm{abs}}$.}
Because the $k_1$ estimator is signed, its running average can hover
near zero even when the token-level teacher$\|$student gap is large in
both directions; this is desirable for gradient estimation but
misleading for monitoring. We therefore log alongside the signed loss
a magnitude-preserving diagnostic
\begin{equation}
\label{eq:abs-loss}
\mathcal{L}_{\mathrm{abs}}(\theta)
\;=\;
\mathbb{E}\!\left[\,\frac{1}{|y|}\sum_{t=1}^{|y|}
\Big|\log\pi_\theta(y_t\mid x,y_{<t}) - \log\pi_{T_k}(y_t\mid x,y_{<t})\Big|\,\right],
\end{equation}
with the outer expectation over $(x,k)\!\sim\!\mathcal{D}$ and
$y\!\sim\!\pi_{\theta_{\mathrm{old}}}$.  $\mathcal{L}_{\mathrm{abs}}$
is computed under \texttt{torch.no\_grad()} and is \emph{not}
back-propagated; it exists purely to give an interpretable, always
non-negative summary of teacher$\|$student proximity that decreases
monotonically as distillation progresses (Figure~\ref{fig:distill-curves}a).

\paragraph{(c) Clipped policy-gradient surrogate $\mathcal{L}_{\mathrm{distill}}$.}
Because rollouts are produced asynchronously by an inference engine at
parameters $\theta_{\mathrm{old}}$ while the training engine has since
advanced to $\theta$, samples are strictly off-policy at the moment of
the gradient step.  Following the standard PPO
correction we treat the negated $k_1$ estimator
as a per-token advantage
\begin{equation}
\hat{A}_t
\;=\;
\operatorname{sg}\!\Big[\,
\log\pi_{T_k}(y_t\mid x,y_{<t})
\;-\;
\log\pi_{\theta_{\mathrm{old}}}(y_t\mid x,y_{<t})
\,\Big],
\end{equation}
compute the (stop-gradient-normalised) importance ratio
\begin{equation}
\rho_t(\theta)
\;=\;
\frac{\pi_\theta(y_t\mid x,y_{<t})}{\operatorname{sg}\!\big[\pi_{\theta_{\mathrm{old}}}(y_t\mid x,y_{<t})\big]},
\end{equation}
and optimize the clipped surrogate
\begin{equation}
\label{eq:distill-loss}
\mathcal{L}_{\mathrm{distill}}(\theta)
\;=\;
-\,\mathbb{E}\!\left[\,\frac{1}{|y|}\sum_{t=1}^{|y|}
\min\!\Big(\rho_t(\theta)\,\hat{A}_t,\;
\mathrm{clip}\!\big(\rho_t(\theta),\,1-\epsilon,\,1+\epsilon\big)\,\hat{A}_t\Big)\right],
\end{equation}
with $\operatorname{sg}[\cdot]$ the stop-gradient operator and
$\epsilon{=}0.2$. Two properties are worth noting. First, when the
inference and training engines are synchronised
($\theta{=}\theta_{\mathrm{old}}$) we have $\rho_t\equiv 1$ and
$\mathcal{L}_{\mathrm{distill}}$ collapses to the on-policy MOPD
gradient of Eq.~\eqref{eq:mopd}. Second, tokens on which the training
engine has already drifted far from the inference engine
($\rho_t$ outside $[1-\epsilon,\,1+\epsilon]$) have their gradient
attenuated by the clip, which we find empirically eliminates the loss
spikes we observed in an unclipped variant during the first few steps
of high-drift domains (tool-agent, deep-search).
$\mathcal{L}_{\mathrm{distill}}$ is the quantity actually
back-propagated in every training step and is plotted in
Figure~\ref{fig:distill-curves}b.

\paragraph{Relation to Eq.~\ref{eq:mopd}.}
Eqs.~\eqref{eq:k1} and~\eqref{eq:distill-loss} together constitute the
sampled, off-policy-corrected estimator of the population objective
stated in Eq.~\eqref{eq:mopd}; Eq.~\eqref{eq:abs-loss} is a
non-optimized diagnostic. In the on-policy limit
($\theta{=}\theta_{\mathrm{old}}$), both
$\mathcal{L}_{\mathrm{abs}}$ (up to the sign convention) and
$\mathcal{L}_{\mathrm{distill}}$ (up to the clip, inactive at
$\rho_t{=}1$) reduce to the token-level reverse KL that
Eq.~\eqref{eq:mopd} minimizes.

\section{Ablations on the Reasoning Expert in MOPD}
\label{app:opd-ablation}

Before scaling MOPD to the full production run, we ran
a progressive teacher-composition study in which teachers were added
one domain at a time. The purpose was not merely to select a final
recipe, but to expose \emph{cross-domain transfer effects} (both
synergies and conflicts) between task-specific experts. The
clearest illustration comes from a pair that differs by a single
teacher: one configuration distills from tool-use, IF, and safety
teachers; the other extends it with a code teacher, holding every
other component fixed. Table~\ref{tab:opd-ablation-code} reports the
two configurations on a subset of evaluation benchmarks.

\begin{table}[h]
\centering
\small
\setlength{\tabcolsep}{4pt}
\caption{Effect of adding a code teacher on top of a
tool-use\,+\,IF\,+\,safety MOPD configuration. All scores are from a
single run under identical evaluation protocols; $\Delta$ is the row
difference. Positive $\Delta$ denotes improvement on every metric. CS-Bench, BS-Bench, LCB-V6 abbreviate Content-SafetyBench,Behavioral-SafetyBench, LiveCodeBench-V6 respectively.}
\label{tab:opd-ablation-code}
\begin{tabular}{lcccccccc}
\toprule
Config & CS-Bench & BS-Bench
       & AIME'25 & AIME'26 &  LCB-V6 & IFEval & IFBench & BFCL \\
\midrule
tool-use + IF + safety   & 96.94 & 76.22 & 90.62 & 91.04 & 78.63 & 92.61 & 81.62 & 76.30 \\
+ code                   & 97.79 & 80.28 & 92.92 & 91.87 & 82.44 & 94.75 & 82.17 & 76.39 \\
\midrule
$\Delta$                 & +0.85 & +4.06 & +2.30 & +0.83 & +3.81 & +2.14 & +0.55 & +0.09 \\
\bottomrule
\end{tabular}
\end{table}

Adding the code teacher produces the following observations.
Reasoning-oriented benchmarks improve substantially:  LiveCodeBench-V6 by $+3.81$,
AIME'25 by $+2.30$, and AIME'26 by $+0.83$. Non-reasoning axes also
improve: Content-SafetyBench (CS-Bench) by $+0.85$,
Behavioral-SafetyBench (BS-Bench) by $+4.06$, IFEval by $+2.14$, and IFBench by $+0.55$. BFCL is essentially unchanged ($+0.09$).

Two lessons emerge, and both directly shape the full
production configuration used in the main results. First, code is a
load-bearing domain rather than a niche one: despite being intended
for code, the code teacher measurably improves mathematical reasoning,
instruction following, and safety metrics simultaneously. Notably, no
mathematics-specific supervision was included at this stage; the
gains on AIME therefore arise purely as cross-domain transfer from
code supervision rather than from direct mathematical training. We
therefore treat code as a first-class domain in every subsequent MOPD
configuration. Second, every capability the practitioner intends to
preserve or improve should be represented by at least one teacher in
the joint distillation. The A/B comparison above shows that adding a
teacher can benefit domains it was not designed to cover, but this
positive transfer is not something to be relied upon in advance:
whether a given cross-domain interaction is synergistic or
competitive is difficult to predict from teacher composition alone,
and capabilities left unrepresented risk drifting. We therefore recommend joint
representation over reliance on cross-domain transfer from adjacent
teachers alone. The two cases above are representative rather than
exhaustive; together they provide the empirical basis for the
final teacher composition adopted in the main experiments.

\end{document}